%% file: main.tex
\documentclass[10pt,twocolumn,letterpaper]{article}

\usepackage[pagenumbers]{cvpr} %

\usepackage{graphicx}
\usepackage{amsmath}
\usepackage{amssymb}
\usepackage{booktabs}
\usepackage[accsupp]{axessibility}  %

\usepackage[pagebackref,breaklinks,colorlinks]{hyperref}

\usepackage[capitalize]{cleveref}

\usepackage{multirow}
\usepackage{bm}
\usepackage{makecell}
\usepackage{float}
\usepackage{enumitem}

\crefname{section}{Sec.}{Secs.}
\Crefname{section}{Section}{Sections}
\Crefname{table}{Table}{Tables}
\crefname{table}{Tab.}{Tabs.}

\def\cvprPaperID{6580} %
\def\confName{CVPR}
\def\confYear{2023}

\begin{document}

\setlength{\abovedisplayskip}{4pt}
\setlength{\belowdisplayskip}{4pt}

\title{High-Fidelity and Freely Controllable Talking Head Video Generation}

\author{Yue Gao \quad Yuan Zhou \quad Jinglu Wang \quad Xiao Li \quad Xiang Ming  \quad Yan Lu \vspace{4pt}\\
    Microsoft Research \\
    {\tt\small \{yuegao, zhouyuan, jinglu.wang, li.xiao, xiangming, yanlu\}@microsoft.com} \\
}

\input{sub_tex/fig_teaser}

\input{subsections/abstract}
\input{subsections/introduction}
\input{subsections/related_works}

\input{subsections/method}

\input{subsections/experiments}
\input{subsections/conclusion}

\clearpage{\thispagestyle{empty}\cleardoublepage}

{\small
\bibliographystyle{ieee_fullname}
\bibliography{main}
}

\end{document}

%% file: sub_tex/fig_teaser.tex
\twocolumn[{%
\renewcommand\twocolumn[1][]{#1}%
\maketitle
\begin{center}
    \centering
    \captionsetup{type=figure}
  \centering
  \begin{minipage}[t]{.5\linewidth}
    \captionsetup[subfigure]{labelformat=empty}
    \centering
    \makebox[6pt]{\raisebox{26pt}{\rotatebox[origin=c]{90}{Input}}}%
    \hspace{0.5mm}
    \subcaptionbox{}
      {\includegraphics[width=0.235\linewidth]{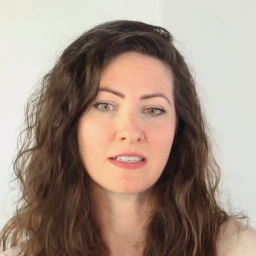}}\hspace{0.5mm}%
      {\includegraphics[width=0.235\linewidth]{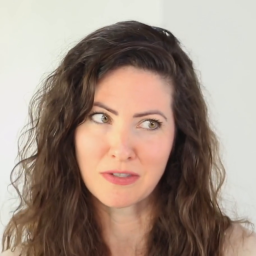}}\hspace{0.5mm}%
      {\includegraphics[width=0.235\linewidth]{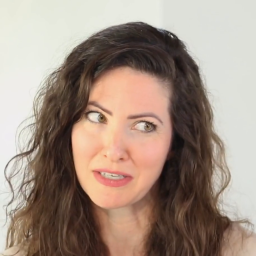}}\hspace{0.5mm}%
      {\includegraphics[width=0.235\linewidth]{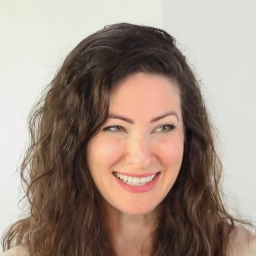}}
    \makebox[6pt]{\raisebox{26pt}{\rotatebox[origin=c]{90}{FOMM}}}%
    \hspace{0.5mm}
    \subcaptionbox{}
      {\includegraphics[width=0.235\linewidth]{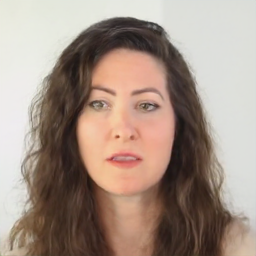}}\hspace{0.5mm}%
      {\includegraphics[width=0.235\linewidth]{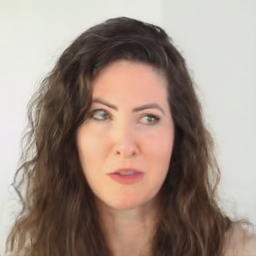}}\hspace{0.5mm}%
      {\includegraphics[width=0.235\linewidth]{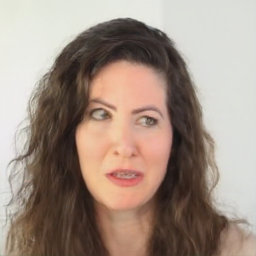}}\hspace{0.5mm}%
      {\includegraphics[width=0.235\linewidth]{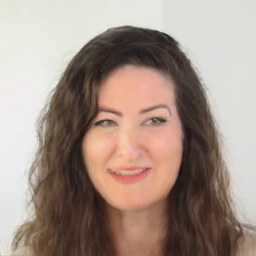}}
    \makebox[6pt]{\raisebox{26pt}{\rotatebox[origin=c]{90}{PECHead}}}%
    \hspace{0.5mm}
    \subcaptionbox{}
      {\includegraphics[width=0.235\linewidth]{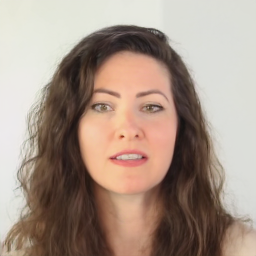}}\hspace{0.5mm}%
      {\includegraphics[width=0.235\linewidth]{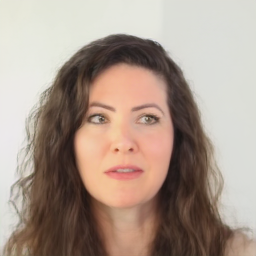}}\hspace{0.5mm}%
      {\includegraphics[width=0.235\linewidth]{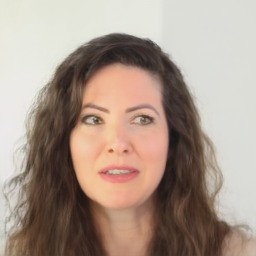}}\hspace{0.5mm}%
      {\includegraphics[width=0.235\linewidth]{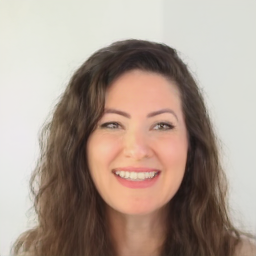}}

    \captionsetup[subfigure]{labelformat=parens,labelsep=period}
    \setcounter{subfigure}{0}
    \subcaption{Frontalization}\label{fig:teaser_front}%
  \end{minipage}%
  \hfill
  \begin{minipage}[t]{.5\linewidth}
    \captionsetup[subfigure]{labelformat=empty}
    \centering
    \subcaptionbox{Input}
      {\includegraphics[width=0.235\linewidth]{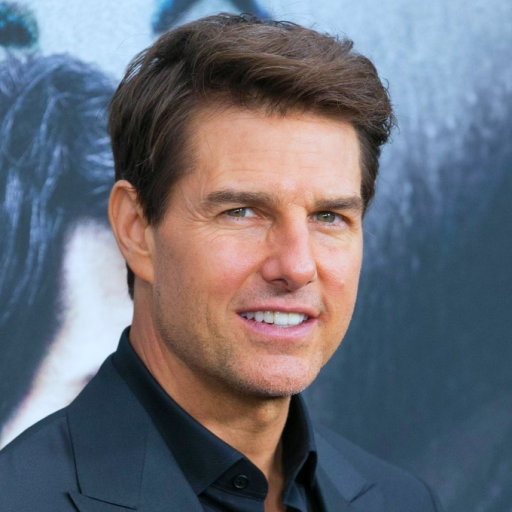}}\hspace{0.5mm}%
    \subcaptionbox{Yaw}
      {\includegraphics[width=0.235\linewidth]{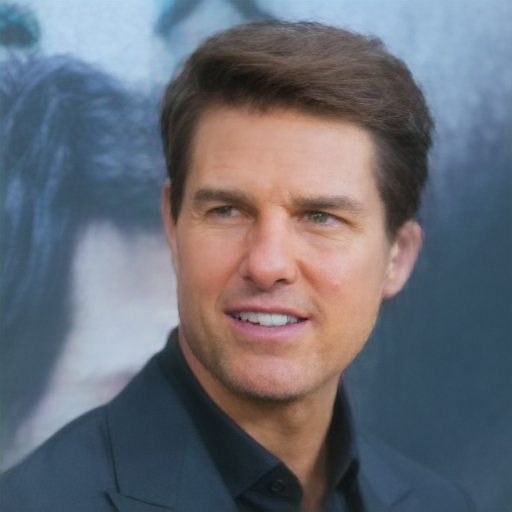}}\hspace{0.5mm}%
    \subcaptionbox{Pitch}
      {\includegraphics[width=0.235\linewidth]{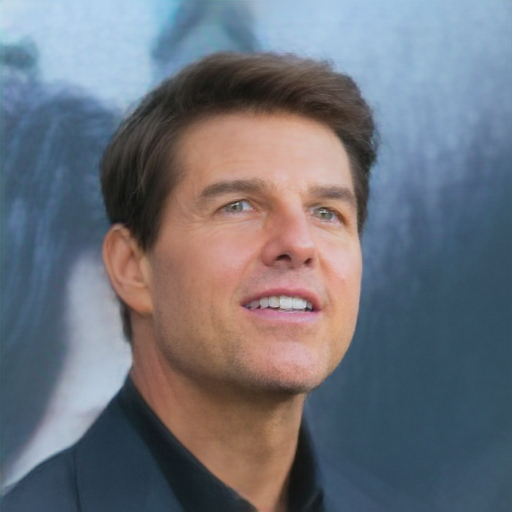}}\hspace{0.5mm}%
    \subcaptionbox{Roll}
      {\includegraphics[width=0.235\linewidth]{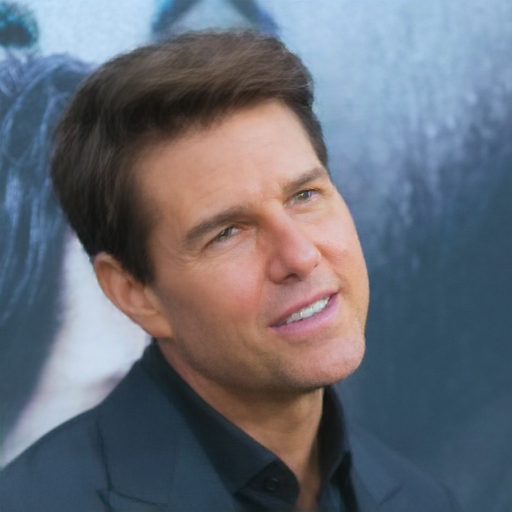}}
    \subcaptionbox{Input}
      {\includegraphics[width=0.235\linewidth]{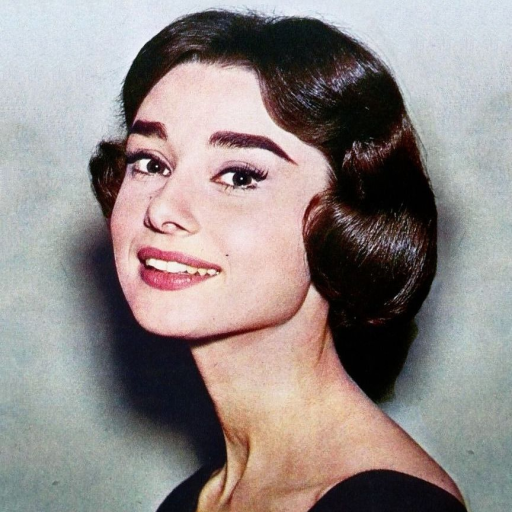}}\hspace{0.5mm}%
    \subcaptionbox{Neutral}
      {\includegraphics[width=0.235\linewidth]{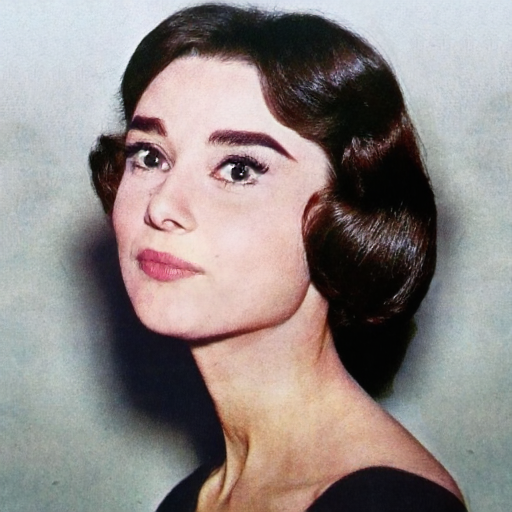}}\hspace{0.5mm}%
    \subcaptionbox{Smile}
      {\includegraphics[width=0.235\linewidth]{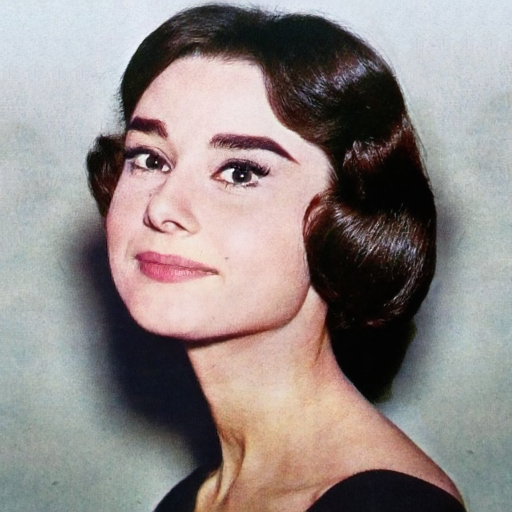}}\hspace{0.5mm}%
    \subcaptionbox{Laugh}
      {\includegraphics[width=0.235\linewidth]{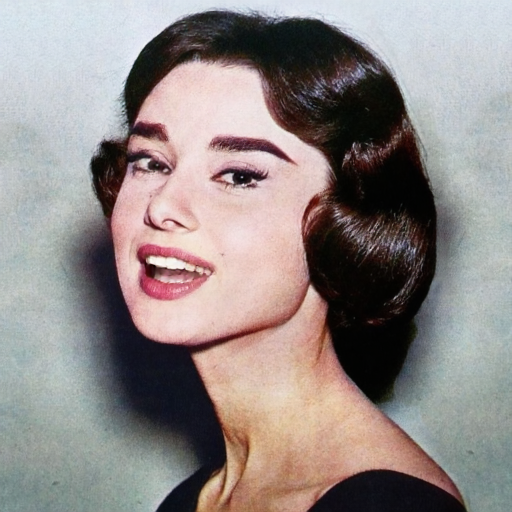}}
    \subcaptionbox{Input}
      {\includegraphics[width=0.235\linewidth]{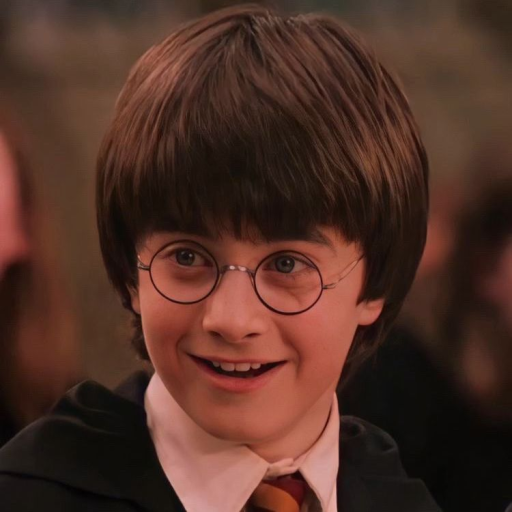}}\hspace{0.5mm}%
    \subcaptionbox{Smile}
      {\includegraphics[width=0.235\linewidth]{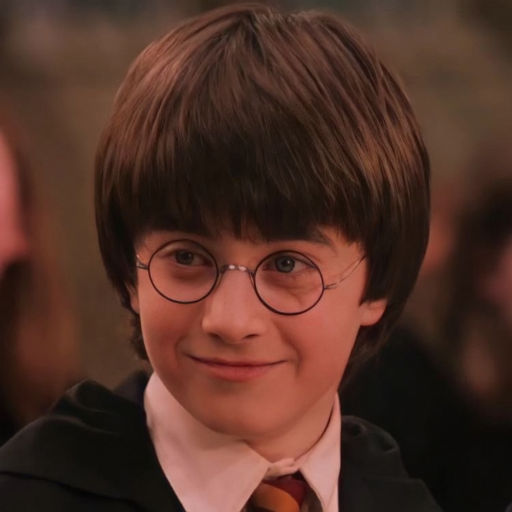}}\hspace{0.5mm}%
    \subcaptionbox{Expressionless}
      {\includegraphics[width=0.235\linewidth]{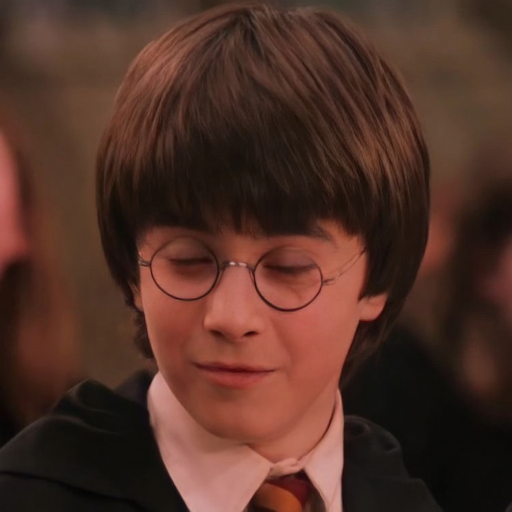}}\hspace{0.5mm}%
    \subcaptionbox{Roll}
      {\includegraphics[width=0.235\linewidth]{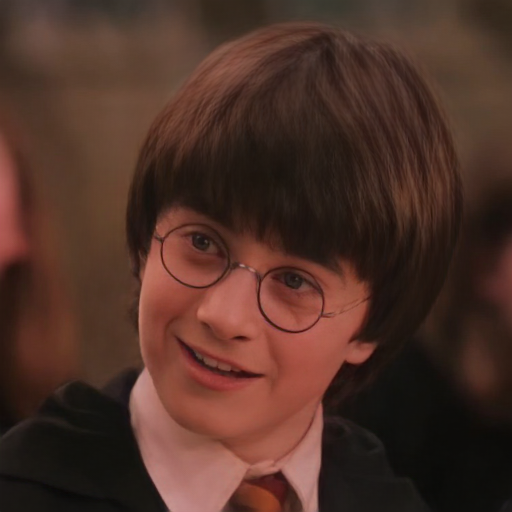}}

    \captionsetup[subfigure]{labelformat=parens,labelsep=period}
    \setcounter{subfigure}{1}
    \subcaption{Free pose and expression editing}\label{fig:teaser_editing}%
  \end{minipage}%
  \caption
    {%
    Presented herein are representative results showcasing the effectiveness of our proposed method in the tasks of frontalization, as well as free pose and expression editing. 
    (a) FOMM~\cite{siarohin2019first} often produces face distortion issues, while our model \textbf{PECHead} generates high-fidelity results.
    (b) Our proposed framework empowers the generation of talking head videos that offer free control over the head pose and expression.
    More results can be found in the supplementary materials.
    \label{fig:teaser}%
    }%
\end{center}%
}]

%% file: subsections/abstract.tex
\begin{abstract}
Talking head generation is to generate video based on a given source identity and target motion.
However, current methods face several challenges that limit the quality and controllability of the generated videos.
First, the generated face often has unexpected deformation and severe distortions.
Second, the driving image does not explicitly disentangle movement-relevant information, such as poses and expressions, which restricts the manipulation of different attributes during generation.
Third, the generated videos tend to have flickering artifacts due to the inconsistency of the extracted landmarks between adjacent frames.
In this paper, 
we propose a novel model that produces high-fidelity talking head videos with free control over head pose and expression.
Our method leverages both self-supervised learned landmarks and 3D face model-based landmarks to model the motion. We also introduce a novel motion-aware multi-scale feature alignment module to effectively transfer the motion without face distortion. Furthermore, we enhance the smoothness of the synthesized talking head videos with a feature context adaptation and propagation module.
We evaluate our model on challenging datasets and demonstrate its state-of-the-art performance.
\end{abstract}

%% file: subsections/introduction.tex
\section{Introduction}
\label{sec:intro}
Talking head video generation is a process of synthesizing a talking head video with a given source identity and target motion.
This process is also called face reenactment when using a driving head to define the relative movement to the source identity~\cite{burkov2020neural}.
This generation technique can be used in various applications, including video conferencing~\cite{wang2021one}, movie effects~\cite{naruniec2020high}, and entertainment~\cite{chan2019everybody}.
Due to the rapid development of deep learning~\cite{he2016deep} and generative adversarial networks (GAN)~\cite{goodfellow2020generative, isola2017image}, impressive works have been conducted on talking head generation~\cite{wang2021one, hong2022depth},
face reenactment~\cite{wu2018reenactgan,zhang2019one,yang2022face,zhang2020freenet, hsu2022dual,bounareli2022finding, nirkin2022fsganv2, yao2021one, thies2016face2face},
and image animation~\cite{siarohin2019first,siarohin2021motion,zhao2022thin,wang2022latent}. These works focus on animating objects beyond the face and head~\cite{siarohin2019animating,siarohin2019first}.

Early works on talking head generation require multiple source and driving images to generate one result~\cite{burkov2020neural, doukas2021headgan}.
Recent works focus on one-shot generation~\cite{yao2021one, zhang2019one,wang2021one}, \ie, using only one source frame to generate the target by transferring the pose information of one driving frame.
Currently, the mainstream works~\cite{siarohin2019animating, siarohin2019first, siarohin2021motion} follow a self-supervised learning pipeline.
They mainly utilized the self-supervised learned landmarks to model the movement of the identity between the source and driving images.
The learned landmarks pairs are first detected from both source and driving images, and then the dense flow field is estimated from the two sets of learned landmarks to transform the source features and guide the reconstruction of the driving image.
To further improve the performance, recent approaches propose to utilize additional information, such as 3D learned landmarks~\cite{wang2021one} and depth map~\cite{hong2022depth}, or enhance the model structure, for example, adopting the flexible thin-plate spline transformation~\cite{duchon1977splines,zhao2022thin}, and representing the motion as a combination of latent vectors~\cite{wang2022latent}.

However, there are still many challenges with these methods.
First, the generated face often has unexpected deformation and severe distortions.
The learned landmarks-based approaches~\cite{siarohin2019first,wang2021one,zhao2022thin}, such as FOMM~\cite{siarohin2019first}, which only utilizes the 2D learned landmarks without face shape constraints, produces frontalization results with apparent face distortions (see Fig.~\ref{fig:teaser_front}).
The predefined landmarks-based methods~\cite{zakharov2019few,yang2022face,hsu2022dual,doukas2021headgan} model the movement between the source and driving images only based on the predefined facial landmarks, leading to the non-facial parts of the head (such as the hair and neck) are not well handled.
Second, all the movement information needs to be obtained via one single driving image.
It is rare and difficult to decouple and manipulate these movement-relevant information, including poses and expressions, when generating the new image.
Third, 
in order to achieve smooth and natural movements in generated videos, prior methods~\cite{siarohin2019first, siarohin2021motion, zhao2022thin} typically incorporate techniques to smoothen the extracted landmarks learned between adjacent frames.
However, the sensitivity and inconsistency of the extracted landmarks pose a challenge in achieving smoothness, resulting in generated videos that are prone to flickering.

To address the above challenges, we propose the \textbf{P}ose and \textbf{E}xpression \textbf{C}ontrollable \textbf{Head} model (\textbf{PECHead}), which can generate high-fidelity video face reenactment results and enable talking head video generation with full control over head pose and expression.
The proposed method first incorporates the learned landmarks and the predefined face landmarks to model the overall head movement and the detailed facial expression changing in parallel.
We utilize the single image-based face reconstruction model~\cite{deng2019accurate} to obtain the face landmarks and project them into 2D image space.
This approach constrains the face to a physically reasonable shape, thereby reducing distortion during motion transfer, as demonstrated in the last row of Fig.~\ref{fig:teaser_front}.
In this work, we introduce the use of learned sparse landmarks for global motion and predefined dense landmarks for local motion, with the Motion-Aware Multi-Scale Feature Alignment (MMFA) module serving to align these two groups of features.
Then we use different coefficients as input conditions to control the estimation of both predefined and learned landmarks, so that we can realize the head pose and expression manipulation (Fig.~\ref{fig:teaser_editing}).
Moreover, inspired by the recent low-level video processing works~\cite{chan2022basicvsr++, li2021deep}, we propose the Context Adaptation and Propagation (CAP) module to further improve the smoothness of the generated video.
Our proposed method is evaluated on multiple talking head datasets, and experimental results indicate that it achieves state-of-the-art performance, generating high-fidelity face reenactment results and talking head videos with the ability to control the desired head pose and facial expression.

Our contributions can be summarized as follows:
\begin{itemize}[itemsep=1pt,topsep=4pt]
\item We propose a novel method, \textbf{PECHead}, that generates high-fidelity face reenactment results and talking head videos.
Our approach leverages head movements to control the estimation of learned and predefined landmarks, enabling free control over the head pose and expression in talking head generation.
\item We incorporate the learned and predefined face landmarks for global and local motion estimation with the proposed Motion-Aware Multi-Scale Feature Alignment module, 
which substantially enhances the quality of synthesized images.
\item We introduce a video-based pipeline with the Context Adaptation and Propagation module to further improve the smoothness and naturalness of the generated videos.
\item Extensive qualitative and quantitative results across several datasets demonstrate the superiority of the proposed framework for high-fidelity video face reenactment and freely controllable talking head generation.
\end{itemize}

%% file: subsections/related_works.tex
\section{Related Works}
\label{sec:related}

\noindent\textbf{Image Animation.}~Image animation is to transfer motion information from one domain to another. Traditional approaches often rely on strong priors such as face meshes~\cite{cao2014displaced}, human keypoints~\cite{chan2019everybody}, or action units~\cite{pumarola2018ganimation,friesen1978facial}.
In recent years, there has been a growing interest in self-supervised methods that only require videos. Monkey-Net~\cite{siarohin2019animating} uses sparse learned landmarks to estimate optical flow for animating arbitrary objects.
FOMM~\cite{siarohin2019first} extends Monkey-Net~\cite{siarohin2019animating} by incorporating local affine transformation.
MRAA~\cite{siarohin2021motion} proposes region-based motion representations, while LIA~\cite{wang2022latent} represents motion as a set of learned motion directions.
These methods eliminate the requirement of explicit structure representations.
Zhao et al.\cite{zhao2022thin} leverage thin-plate spline transformation\cite{duchon1977splines} for motion estimation, which is more flexible than traditional approaches.

\noindent\textbf{Talking Head Generation.}~In recent years, a significant amount of research has been dedicated to the task of face reenactment or talking head generation.
This is largely owing to the development of large-scale face data~\cite{huang2008labeled, yang2016wider}, 3D morphable face model (3DMM) and 3D mesh~\cite{gerig2018morphable, li2017learning}, neural radiance fields (NeRF)\cite{mildenhall2021nerf}, and face landmark detectors\cite{wu2018look, lv2017deep}.
We will discuss these methods in details as follows.

3D face model-based methods.
For example, Face2Face\cite{thies2016face2face} employs a deformation transfer approach to track facial expressions of both source and driving videos, followed by re-rendering of the synthesized faces.
Ma et al.~\cite{ma2019real} reconstructs an individual-specific face model with high-resolution facial geometry and appearance.

Direct synthesis-based models synthesize target faces by decoding latent appearance and motion representations.
For instance, Zakharov et al.\cite{zakharov2019few} introduce the first direct synthesis method for face reenactment.
LPD\cite{burkov2020neural} utilizes head pose augmentation, while DAE-GAN~\cite{zeng2020realistic} disentangles identity and pose representations using the deforming autoencoder~\cite{shu2018deforming}.

3D mesh-based methods utilize neural head models to synthesize realistic head avatars from videos.
Grassal et al.~\cite{grassal2022neural} proposes a neural head model that provides a disentangled shape and appearance representation.
ROME~\cite{khakhulin2022realistic} uses a single image to estimate a person-specific head mesh and texture to synthesize neural head avatars.

NeRF-based methods use NeRF as a novel 3D proxy to represent the head geometry and appearance.
For example, AD-NeRF~\cite{guo2021ad} proposes using NeRF for audio-driving talking head video generation.
Head-NeRF~\cite{hong2022headnerf} uses NeRF to control the pose and various semantic attributes of the generated images.
However, NeRF-based methods are not effective in generalizing across identities, and the models are relatively complex compared to the sparse landmark-based models.

Warping-based methods use learned landmarks/regions pairs to estimate motion fields~\cite{siarohin2019animating,siarohin2019first,siarohin2021motion}, performs warping on the feature maps, and generates images. X2Face~\cite{wiles2018x2face} uses latent vectors that are learned to be predictable of warping fields. Bi-layer~\cite{zakharov2020fast} employs a bi-layer representation via summing two components, a coarse image directly predicted by a rendering network and a warped texture image. PIRender~\cite{ren2021pirenderer} controls the face motions directly with 3DMMs. OSFV~\cite{wang2021one} extracts 3D learned landmarks with 3D convolution nets for better modeling the head, and utilizes the rotation matrix to transform the overall viewpoint but not for free control of all head poses and expressions. HeadGAN~\cite{doukas2021headgan} and Face2Face$^\rho$\cite{yang2022face} are two such methods that estimate motion information from input images using 3D meshes and landmarks, respectively. DaGAN\cite{hong2022depth} presents a self-supervised depth estimator and cross-modal attention to generate motion fields. While these methods have shown promising results, there is still room for improvement in terms of flexibility, physics-consistency, and video smoothness. To address these limitations, this paper proposes a novel approach that leverages both self-supervised learned landmarks and predefined landmarks for motion transfer while also improving the smoothness of the resulting videos.

%% file: subsections/method.tex
\section{Method}
\label{sec:method}

\input{sub_tex/fig_framework}

\subsection{Overview}
\label{sec:overview}

This section describes the proposed method \textbf{PECHead},
(see Fig.~\ref{fig:framework} for illustration), which mainly contains four parts: Generator $G$, Face Shape Reconstructor $R$, Head Pose-Aware Keypoint Estimator $E$, and Multi-Scale Discriminator $D$.
Our framework follows the basic pipeline proposed by Siarohin et al.~\cite{siarohin2019animating,siarohin2019first,siarohin2021motion}.
We first extract the face coefficients and predefined landmarks through $R$, and then estimate the learned landmarks through $E$ with the head pose and expression as conditions.
The generator $G$ takes the predefined and learned landmarks pairs to estimate the dense flow and generates the results.
During training, our model takes two sequences with the same subject and number of adjacent frames.
We denote the frames in these two sequences as source frame $x^s_t \in \mathbb{R} ^ {3{\times} H {\times} W}$ and driving frame $x^d_t \in \mathbb{R} ^ {3{\times} H {\times} W} $, where $1 {\le} t {\le} T$, $T$ is the sequence length, and $H {\times} W$ is the spatial size.
The model is learned to reconstruct the driving frame $x^d_t$, and the synthesized frame is denoted as $\hat{x}^d_t$.
In the following sections, the frame index $t{-}1$ or $t$ are omitted for brevity, except when necessary.
At test time, %
we can extract the coefficients from the driving frames or modify the coefficients of the source frames, to get different landmarks pairs.
This allows us to transfer the motion from the driving frames or edit the source frames.

\noindent\textbf{Generator.}~Generator $G$ mainly contains the encoder, bottleneck module, and decoder.
The encoder extracts the raw feature $f^r$ of the current source frame.
The bottleneck module aligns the raw feature $f^r$ to the driving frame and adapts it to the context information.
The details will be discussed in Sec.~\ref{sec:mmfa} and Sec.~\ref{sec:cap}.
The decoder generates the reconstructed frames $\hat{x}^d$ based on the adapted feature $f^c$.

\noindent\textbf{Face Shape Reconstructor.}~As mentioned in Sec.~\ref{sec:intro}, existing self-supervised learned landmarks-based models can not freely control the head poses and facial expressions.
To solve this problem, we incorporate the predefined face landmarks~\cite{bulat2017far} with the learned landmarks.
Specifically, a single image-based Face Shape Reconstructor $R$ is adopted to extract the landmarks $l^s, l^d$, head pose $p^s, p^d$ and expression $e^s, e^d$ from the source and driving frames, respectively,%
\begin{equation}
\begin{aligned}
l^z, p^z, e^z = R(x^z); z \in \{s, d\}.%
\end{aligned}
\end{equation}
Our Reconstructor $R$ is derived from the state-of-the-art face reconstruction model~\cite{deng2019accurate}, which uses a ResNet~\cite{he2016deep} to obtain a set of coefficients and fits a Basel Face Model (BFM)~\cite{blanz1999morphable, gerig2018morphable}.
We can further compute the 3D landmarks and project them to the 2D space.
When calculating the driving landmarks $l^d$, the other coefficients (\ie, identity, texture, and lighting) extracted from the source frame are used to preserve the identity.

\noindent\textbf{Head Pose-Aware Keypoint Estimator.}~Existing keypoint-based models~\cite{siarohin2019first, zhao2022thin, hong2022depth, wang2021one} directly feed the source and driving frames to keypoint detector to obtain the learned landmarks pairs.
Instead, we use the source frame $x^s$ conditioned with corresponding head pose $p$ and expression $e$ to obtain source learned landmarks $k^s$ and driving learned landmarks $k^d$, facilitating manually specified head pose and expression editing.
This process is formulated as,
\begin{equation}
    k^z = E(x^s, p^z, e^z); z \in \{s, d\}, k \in \mathbb{R} ^ {K \times 2},
\end{equation}
where $K$ is the number of learned landmarks, we set $K{=}10$.
The $E$ is trained to detect the learned landmarks based on the appearance provided from the source frame $x^s$ obeying the head pose $p$ and expression $e$.
The head pose $p$ and expression $e$ are injected with AdaIN~\cite{huang2017arbitrary} module.
As the coefficients of the face model are decoupled by definition, the different learned and predefined landmarks $k', l'$ can be obtained by modifying the head pose $p$ or expression $e$, and manipulated frame $\hat{x}'$ can be generated correspondingly.

\noindent\textbf{Multi-Scale Discriminator.}~Following the existing generative models~\cite{isola2017image, karras2019style, karras2020analyzing, park2019semantic, wang2021one, siarohin2019first, hong2022depth}, we utilize a Multi-scale Patch Discriminator $D$ to encourage the generator $G$ produce more realistic frames.

\input{sub_tex/fig_mmfa}

\subsection{Motion-Aware Multi-Scale Feature Alignment}
\label{sec:mmfa}

Although both the learned and predefined landmarks are represented in 2D image space, our experimental results demonstrate that directly merging these points in series does not give satisfactory results.
As the learned ones are freely learned by the model, while the predefined ones are artificially defined, their physical meanings of them are different.
Therefore, we propose the Motion-Aware Multi-Scale Feature Alignment (MMFA) module to incorporate the learned sparse and predefined dense landmarks.

As shown in Fig.~\ref{fig:mmfa}, the MMFA correlates the predefined landmarks $l^s, l^d$ and the learned landmarks $k^s, k^d$ for deforming the raw feature extracted by the encoder.
The sparse learned landmarks detected from the whole frame can provide more global motion information, \ie, the overall head movement.
And the landmarks can be used for modeling more details of the motion, such as expression changing, as they are estimated from the face shape model.
We use two motion estimators $\Phi$ to estimate the global and local motion information based on the learned and predefined landmarks, respectively.
Following FOMM~\cite{siarohin2019first}, the Gaussian heatmap-based motion estimators are employed.
The global motion is applied on the downsampled raw feature $f^r_{\text{low}} \in \mathbb{R} ^{C^1 \times h/2 \times w/2}$, which contains higher level information.
And the local motion is applied on the raw feature $f^r \in \mathbb{R} ^{C^2 \times h \times w}$ for more details, where $C^1$ and $C^2$ are the channel dimension, $h {\times} w$ is the spatial size of the feature map.
The motion information contains two parts, deformation $w^{s \to d} \in \mathbb{R} ^{2 \times h \times w}$, and occlusion $o^{s \to d} {\in} \mathbb{R} ^{h \times w}$ (omitted in Fig.~\ref{fig:mmfa} for brevity).
With the deformation and occlusion map, the raw feature can be warped as $f^w = \mathcal{W}(f^r, w^{s \to d}) \cdot o^{s \to d} + ( 1 - o^{s \to d}) \cdot f^r $, where $\mathcal{W}$ is the warping operation.%

\input{sub_tex/fig_dcna}

According to Chan et al.~\cite{chan2021understanding, chan2022basicvsr++}, the deformable alignment~\cite{ke2018sparse} demonstrates significant improvements over the flow-based alignment.
As shown in Fig.~\ref{fig:dcna}, the deformable alignment takes the last feature map $f^m$ in the motion estimator and deformation map $w^{s\to d}$ to compute offsets $\delta^{s\to d}$ and masks $\epsilon^{s\to d}$ of the deformable convolution (DCN)~\cite{dai2017deformable}, and then a DCN is applied,
\begin{equation}
\begin{aligned}
    \delta^{s\to d} &= w^{s\to d} + \mathcal{C}^o(c(f^m, w^{s\to d})), \\
    \epsilon^{s\to d} &= \sigma(\mathcal{C}^m(c(f^m, w^{s\to d})), \\
    f^d &= \mathcal{D}(f^r; \delta^{s \to d}, \epsilon^{s\to d}),
\end{aligned}
\end{equation}
where $c$ is the concatenation operation, $\mathcal{C}^{\{o, m\}}$ denotes convolution blocks, $\sigma$ is the sigmoid and $\mathcal{D}$ stands for the DCN.
We apply the alignment on both global and local levels.

\subsection{Context Adaptation and Propagation}

We introduce the Context Adaptation and Propagation (CAP) module to make the model produce smooth videos.
The illustration of CAP is shown in Fig.~\ref{fig:cap}.
First, the raw feature $f^r_{t-1}$ of the previous source frame is sequentially warped with the frame flow $w_{t-1 \to t}$ and the current local deformation $w^{s \to d}_{\text{local}}$ (Two-Step Warping). The frame flow is computed on $x^s_{t-1}$ and $x^s_t$ using image-based flow estimator~\cite{teed2020raft}.
Second, the warped previous raw feature $\tilde{f}^r_{t-1}$, hidden feature $h_{t-1}$, adapted feature $f^c_{t-1}$ and the current aligned feature $f^a_t$ are concatenated.
We further feed the concatenated feature to the Context Adaptation module, which is composed of several convolution blocks, to get the feature in the same spatial and channel size with $f^a_t$.
Then, the feature is further refined with the Feature Refinement module.
After that, we get the adapted feature $f^c_t$ for the current frame, and the hidden feature is updated with the adapted feature $f^c_t$ using a ConvGRU block~\cite{cho2014properties}. %
The proposed CAP module can be formed as,
\begin{equation}
    \begin{aligned}
        h_t &= \text{ConvGRU}(f^c_{t-1}, h_{t-1}), \\
        \tilde{f}^r_{t-1} &= \mathcal{W}(\mathcal{W}(f^r_{t-1}, w_{t-1 \to t}), w^{s \to d}_{\text{local}}), \\
        f^c_t &= \text{FR}(\text{CA}(c(f^a_t, \tilde{f}^r_{t-1}, f^c_{t-1}, h_{t-1}))),
    \end{aligned}
\end{equation}
where $\text{CA}, \text{FR}$ represent the Context Adaptation and Feature Refinement submodules, respectively.
The hidden feature $h$ and the previous adapted feature $f^c$ are initialized to zeros for the first frame.
\input{sub_tex/fig_cap}

\input{sub_tex/fig_recon}

\input{sub_tex/tab_recon}

\subsection{Objective Function}
Following existing works~\cite{siarohin2019first, siarohin2021motion}, our model is trained with the reconstruction task.
We briefly discuss these losses and leave the details in the supplementary material.

\noindent\textbf{Pixel-wise Loss $\mathcal{L}_{p}$.}~The pixel-wise loss is employed to ensure the synthesis frames are similar to the driving frames.

\noindent\textbf{Perceptual Loss $\mathcal{L}_{v}$.}~Similar to existing methods~\cite{siarohin2019first, siarohin2021motion, zhao2022thin, wang2021one}, we use a pre-trained VGG~\cite{johnson2016perceptual} to guarantee consistency of high level characteristics between driving frame $x^d$ and generated frame $\hat{x}^d$.

\noindent\textbf{Learned Landmarks Loss $\mathcal{L}_{k}$.}~The learned landmarks loss~\cite{wang2021one} is used to encourage the estimated learned landmarks $k$ to spread out across the whole frame.

\noindent\textbf{Equivariance Loss $\mathcal{L}_{e}$.}~The equivariance loss~\cite{siarohin2019first,siarohin2021motion} is applied to constrain the consistency of Head Pose-Aware Keypoint Estimator $E$.

\noindent\textbf{Warping Loss $\mathcal{L}_{w}$.}~This loss is designed to ensure the motion estimators to predict the deformations reasonably, making the warped source frame closer to the driving frame.

\noindent\textbf{GAN Loss $\mathcal{L}_{G}, \mathcal{L}_{D}$.}~We adopt the hinge loss as our adversarial loss~\cite{lim2017geometric}, and two different scale patch discriminator is used for better performance~\cite{isola2017image}.

\noindent\textbf{Full Objective Function.}~The total loss of the generation step is formulated as,
\begin{equation}
    L_G = \lambda_{p} \mathcal{L}_{p} + \lambda_{v} \mathcal{L}_{v} + \lambda_{k} \mathcal{L}_{k} + \lambda_{e} \mathcal{L}_{e} + \lambda_{w} \mathcal{L}_{w} + \lambda_{G} \mathcal{L}_{G},
\end{equation}
where $\lambda_{p}, \lambda_{v}, \lambda_{k}, \lambda_{e}, \lambda_{w} $ and $\lambda_{G}$ are the weights of loss functions.
And the loss of the discrimination step is formulated as $L_D = \mathcal{L}_{D}$.
We follow the standard GAN practice~\cite{isola2017image} to train the model.

%% file: sub_tex/fig_framework.tex
\begin{figure*}
    \centering
    \includegraphics[width=\textwidth]{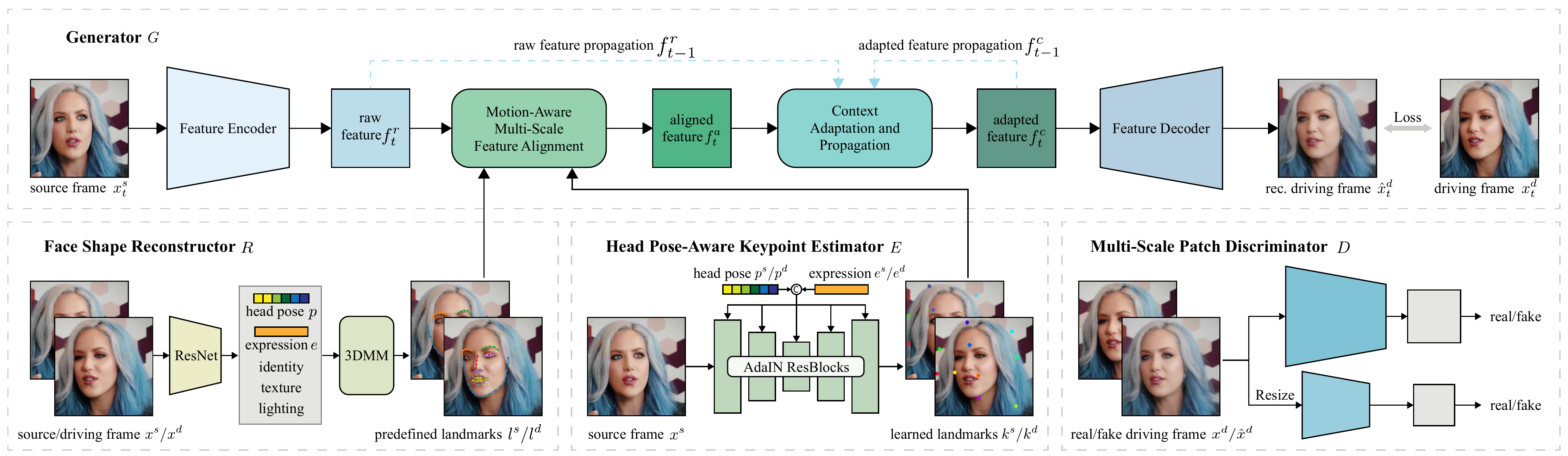}
    \vspace{-5.5mm}
    \caption{The overview of our method, which contains four parts:
    a) the Generator $G$;
    b) the Face Shape Reconstructor $R$;
    c) the Head Pose-Aware Keypoint Estimator $E$;
    and d) the Multi-Scale Discriminator $D$.
    The light blue dash arrows stand for the feature propagation.
    }
    \label{fig:framework}
    \vspace{-4mm}
\end{figure*}

%% file: sub_tex/fig_mmfa.tex
\begin{figure}
    \centering
    \includegraphics[width=\linewidth]{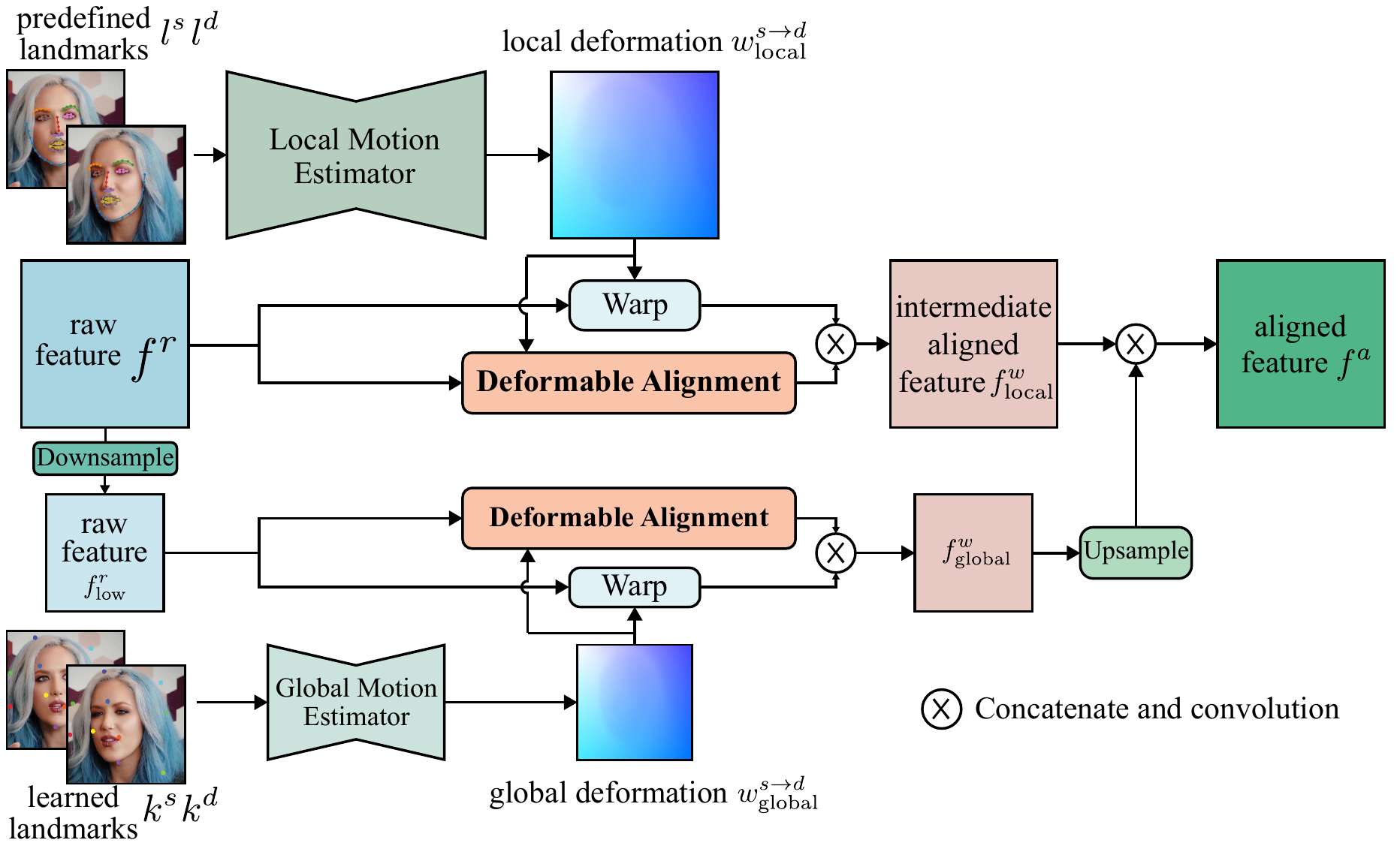}
    \vspace{-6.5mm}
    \caption{Motion-Aware Multi-Scale Feature Alignment module.}
    \label{fig:mmfa}
    \vspace{-5.5mm}
\end{figure}

%% file: sub_tex/fig_dcna.tex
\begin{figure}
    \centering
    \includegraphics[width=\linewidth]{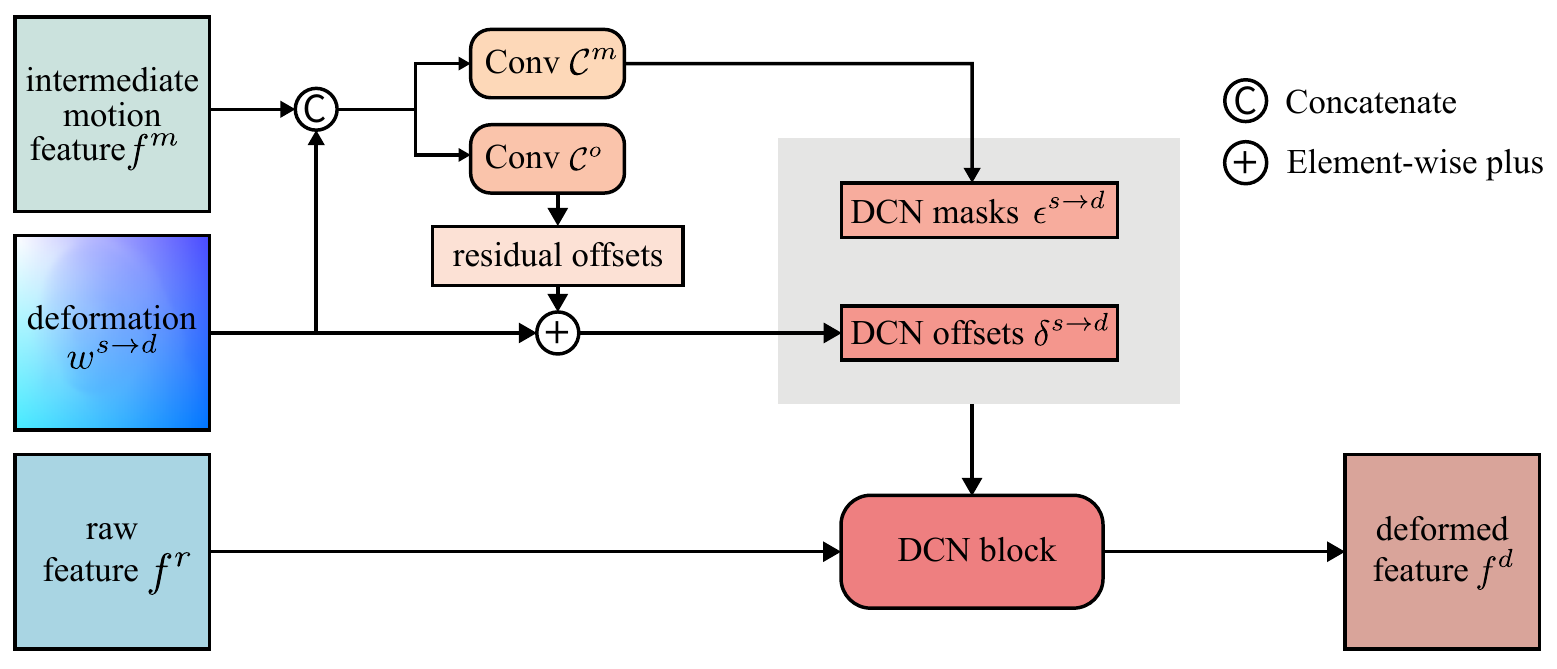}
    \vspace{-7.5mm}
    \caption{Deformable Alignment module.}
    \label{fig:dcna}
    \vspace{-5.5mm}
\end{figure}

%% file: sub_tex/fig_cap.tex
\label{sec:cap}
\begin{figure}
    \centering
    \includegraphics[width=\linewidth]{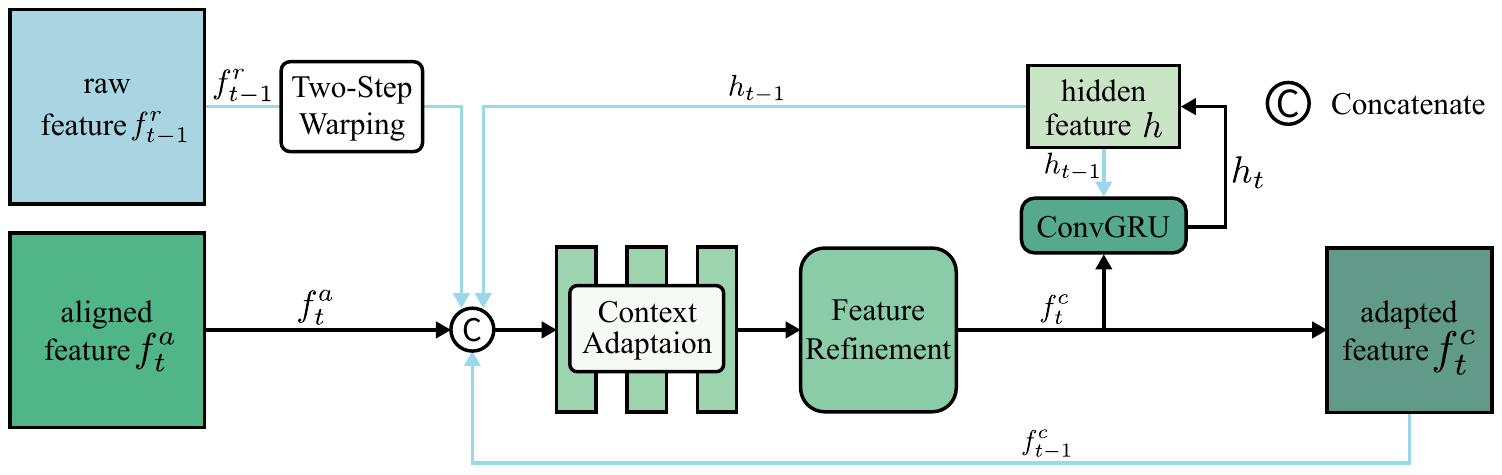}
    \vspace{-6.5mm}
    \caption{Context Adaptation and Propagation module.
    }
    \label{fig:cap}
    \vspace{-5.5mm}
\end{figure}

%% file: sub_tex/fig_recon.tex
\begin{figure*}
    \captionsetup[subfigure]{aboveskip=1pt} %
    \captionsetup[subfigure]{labelformat=empty}
    \captionsetup[subfigure]{font={scriptsize}}
    \centering
    \begin{subfigure}[t]{0.097\linewidth}
        \includegraphics[width=\linewidth]
        {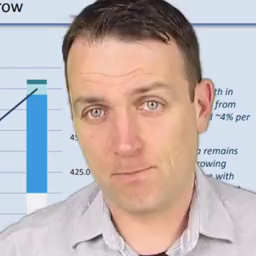} \\
        \vspace{-4.6mm}
        \includegraphics[width=\linewidth]
        {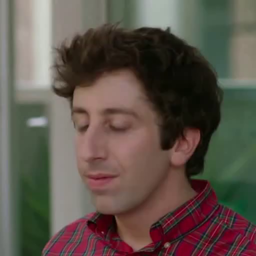} \\
        \vspace{-4.6mm}
        \includegraphics[width=\linewidth]
        {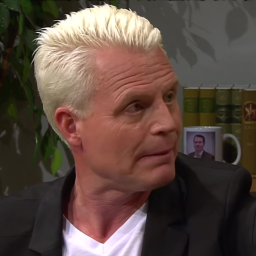} \\
        \vspace{-4.6mm}
        \includegraphics[width=\linewidth]
        {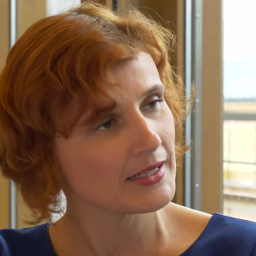} \\
        \vspace{-4.6mm}
        \includegraphics[width=\linewidth]
        {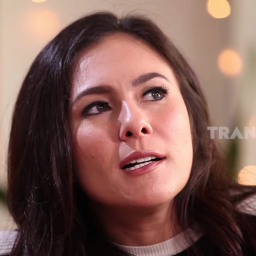}
        \caption{Source}
    \end{subfigure}%
    \begin{subfigure}[t]{0.097\linewidth}
        \includegraphics[width=\linewidth]
        {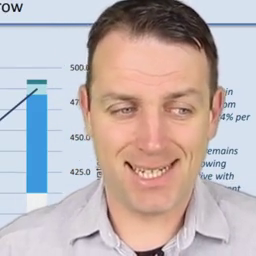} \\
        \vspace{-4.6mm}
        \includegraphics[width=\linewidth]
        {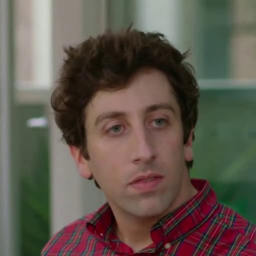} \\
        \vspace{-4.6mm}
        \includegraphics[width=\linewidth]
        {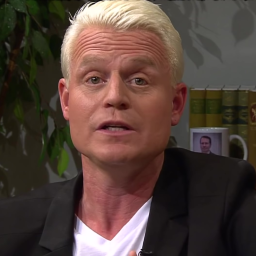} \\
        \vspace{-4.6mm}
        \includegraphics[width=\linewidth]
        {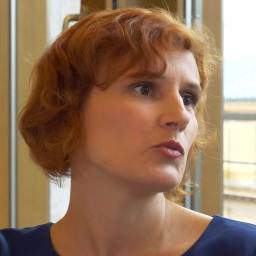} \\
        \vspace{-4.6mm}
        \includegraphics[width=\linewidth]
        {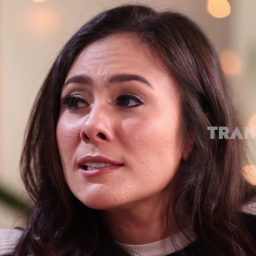}
        \caption{Driving}
    \end{subfigure}\hspace{1mm}%
    \begin{subfigure}[t]{0.097\linewidth}
        \includegraphics[width=\linewidth]
        {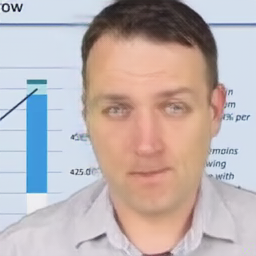} \\
        \vspace{-4.6mm}
        \includegraphics[width=\linewidth]
        {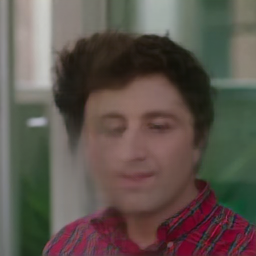} \\
        \vspace{-4.6mm}
        \includegraphics[width=\linewidth]
        {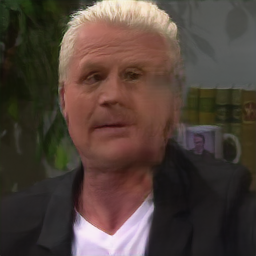} \\
        \vspace{-4.6mm}
        \includegraphics[width=\linewidth]
        {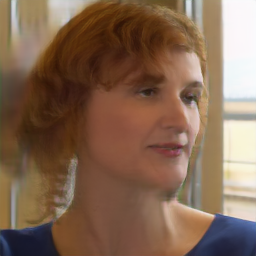} \\
        \vspace{-4.6mm}
        \includegraphics[width=\linewidth]
        {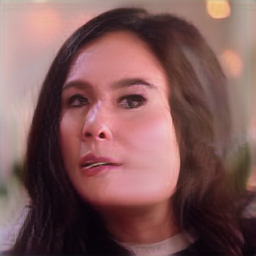}
        \caption{FOMM~\cite{siarohin2019first}}
    \end{subfigure}%
    \begin{subfigure}[t]{0.097\linewidth}
        \includegraphics[width=\linewidth]
        {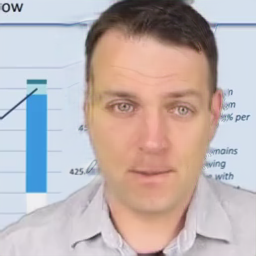} \\
        \vspace{-4.6mm}
        \includegraphics[width=\linewidth]
        {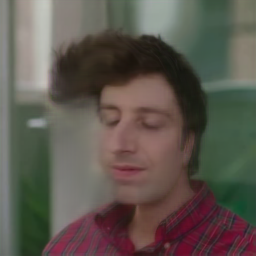} \\
        \vspace{-4.6mm}
        \includegraphics[width=\linewidth]
        {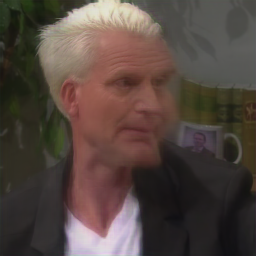} \\
        \vspace{-4.6mm}
        \includegraphics[width=\linewidth]
        {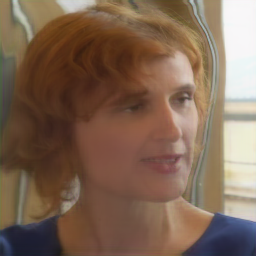} \\
        \vspace{-4.6mm}
        \includegraphics[width=\linewidth]
        {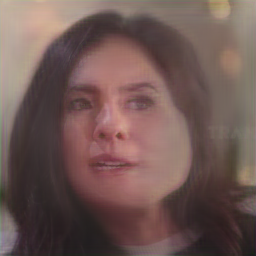}
        \caption{MRAA~\cite{siarohin2021motion}}
    \end{subfigure}%
    \begin{subfigure}[t]{0.097\linewidth}
        \includegraphics[width=\linewidth]
        {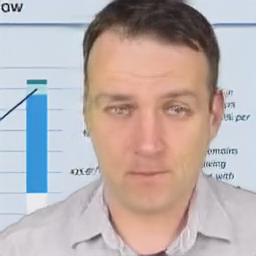} \\
        \vspace{-4.6mm}
        \includegraphics[width=\linewidth]
        {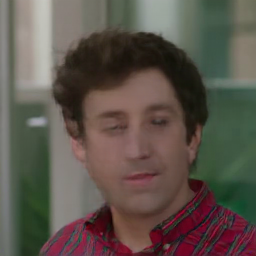} \\
        \vspace{-4.6mm}
        \includegraphics[width=\linewidth]
        {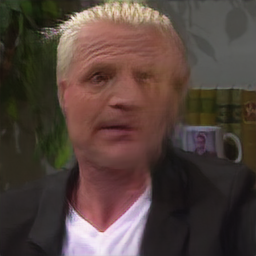} \\
        \vspace{-4.6mm}
        \includegraphics[width=\linewidth]
        {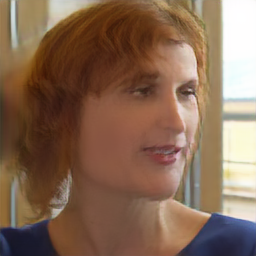} \\
        \vspace{-4.6mm}
        \includegraphics[width=\linewidth]
        {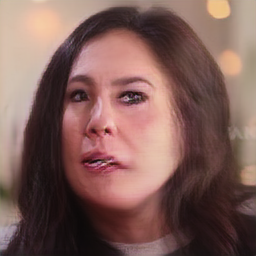}
        \caption{OSFV~\cite{wang2021one}}
    \end{subfigure}%
    \begin{subfigure}[t]{0.097\linewidth}
        \includegraphics[width=\linewidth]
        {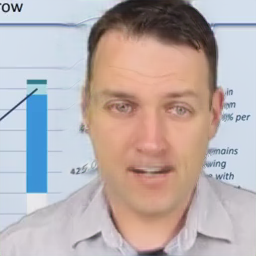} \\
        \vspace{-4.6mm}
        \includegraphics[width=\linewidth]
        {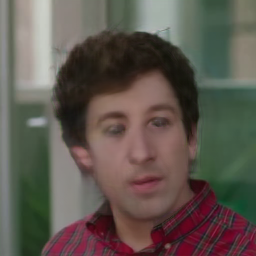} \\
        \vspace{-4.6mm}
        \includegraphics[width=\linewidth]
        {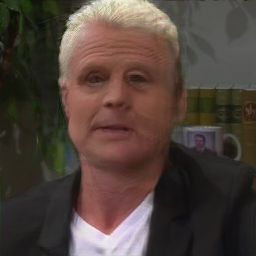} \\
        \vspace{-4.6mm}
        \includegraphics[width=\linewidth]
        {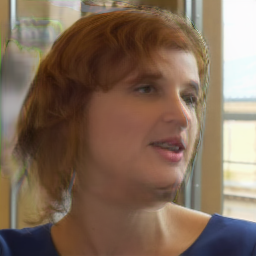} \\
        \vspace{-4.6mm}
        \includegraphics[width=\linewidth]
        {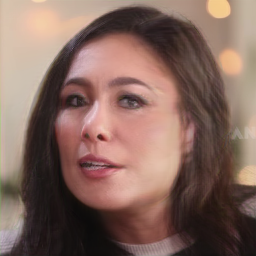}
        \caption{TPSMM~\cite{zhao2022thin}}
    \end{subfigure}%
    \begin{subfigure}[t]{0.097\linewidth}
        \includegraphics[width=\linewidth]
        {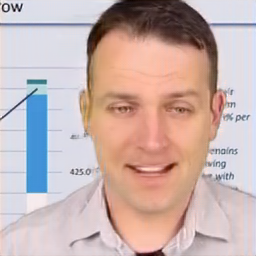} \\
        \vspace{-4.6mm}
        \includegraphics[width=\linewidth]
        {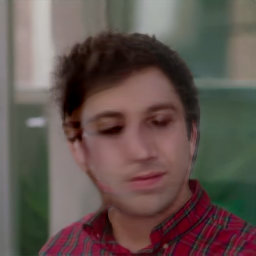} \\
        \vspace{-4.6mm}
        \includegraphics[width=\linewidth]
        {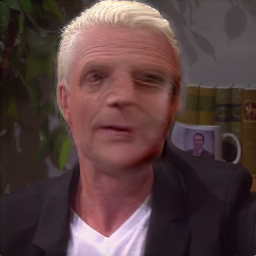} \\
        \vspace{-4.6mm}
        \includegraphics[width=\linewidth]
        {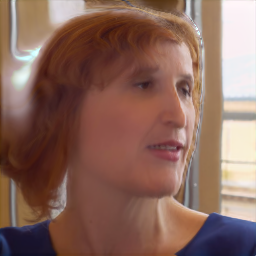} \\
        \vspace{-4.6mm}
        \includegraphics[width=\linewidth]
        {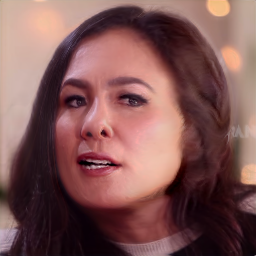}
        \caption{LIA~\cite{wang2022latent}}
    \end{subfigure}%
    \begin{subfigure}[t]{0.097\linewidth}
        \includegraphics[width=\linewidth]
        {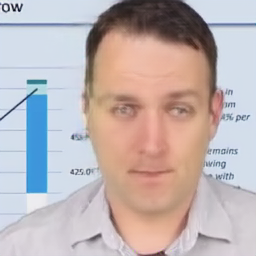} \\
        \vspace{-4.6mm}
        \includegraphics[width=\linewidth]
        {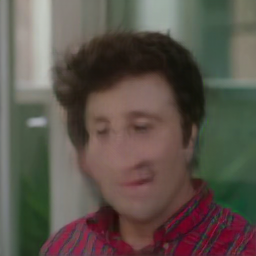} \\
        \vspace{-4.6mm}
        \includegraphics[width=\linewidth]
        {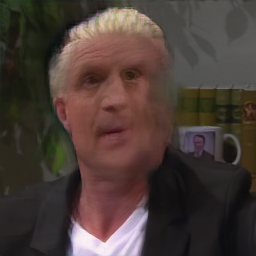} \\
        \vspace{-4.6mm}
        \includegraphics[width=\linewidth]
        {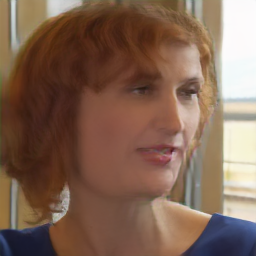} \\
        \vspace{-4.6mm}
        \includegraphics[width=\linewidth]
        {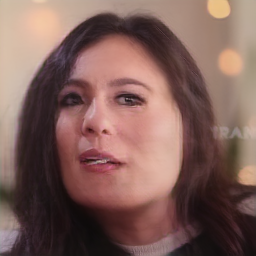}
        \caption{DaGAN\cite{hong2022depth}}
    \end{subfigure}%
    \begin{subfigure}[t]{0.097\linewidth}
        \includegraphics[width=\linewidth]
        {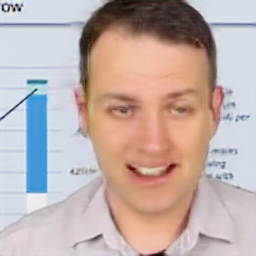} \\
        \vspace{-4.6mm}
        \includegraphics[width=\linewidth]
        {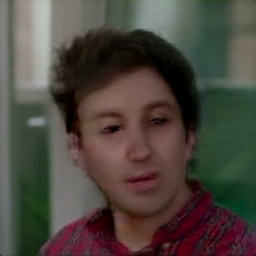} \\
        \vspace{-4.6mm}
        \includegraphics[width=\linewidth]
        {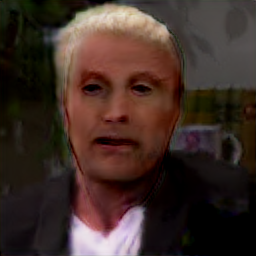} \\
        \vspace{-4.6mm}
        \includegraphics[width=\linewidth]
        {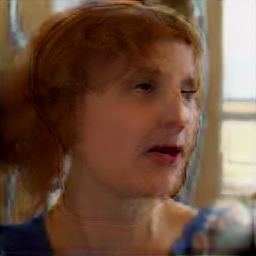} \\
        \vspace{-4.6mm}
        \includegraphics[width=\linewidth]
        {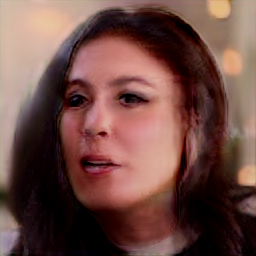}
        \caption{Face2Face$^\rho$~\cite{yang2022face}}
    \end{subfigure}%
    \begin{subfigure}[t]{0.097\linewidth}
        \includegraphics[width=\linewidth]
        {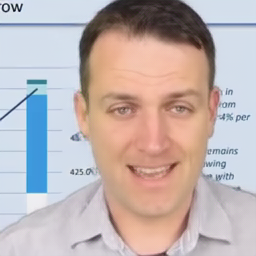} \\
        \vspace{-4.6mm}
        \includegraphics[width=\linewidth]
        {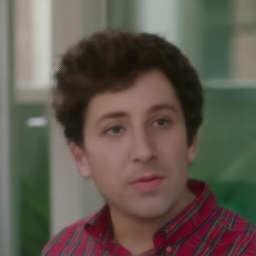} \\
        \vspace{-4.6mm}
        \includegraphics[width=\linewidth]
        {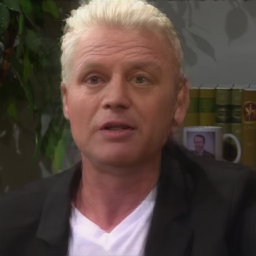} \\
        \vspace{-4.6mm}
        \includegraphics[width=\linewidth]
        {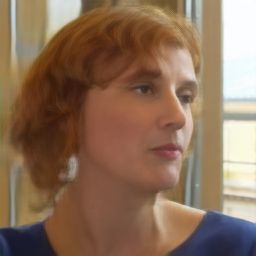} \\
        \vspace{-4.6mm}
        \includegraphics[width=\linewidth]
        {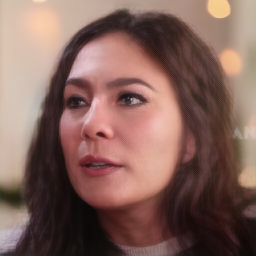}
        \caption{\textbf{PECHead}}
    \end{subfigure}
    \vspace{-3mm}
    \caption{Comparison of same-identity video reconstruction results obtained by the proposed method and other state-of-the-art approaches.}
    \label{fig:sivr}
    \vspace{-2mm}
\end{figure*}

%% file: sub_tex/tab_recon.tex
\begin{table*}[t]
    \caption{Quantitative results of different methods on four datasets for the same-identity video reconstruction.}
    \label{tab:sivr}
    \vspace{-3mm}
    \centering
    \resizebox{\textwidth}{!}{%
    \begin{tabular}{l|ccccc|ccccc|ccccc|ccccc}
    \toprule
    \multirow{2}{*}{Methods} & \multicolumn{5}{c|}{VoxCeleb2} & \multicolumn{5}{c|}{TalkHead-1KH} & \multicolumn{5}{c|}{CelebV-HQ} & \multicolumn{5}{c}{VFHQ} \\
                                         & $L_1$       & MS-SSIM      & PSNR         & FID          & AKD          &$L_1$ & MS-SSIM     & PSNR         & FID          & AKD          &$L_1$& MS-SSIM      & PSNR         & FID          & AKD          &$L_1$& MS-SSIM      & PSNR         & FID         & AKD \\
    \midrule
    FOMM~\cite{siarohin2019first}        &        0.0481 &        0.838 &        23.02 &        25.90 &        1.219 &        0.0431 &        0.821 &        23.28 &        33.22 &        2.905 &        0.0602 &        0.769 &        21.85 &        62.84 &        3.453 &        0.0526 &        0.780 &        21.76 &        47.82 &        2.868 \\
    MRAA~\cite{siarohin2021motion}       &        0.0353 &        0.881 &        25.94 &        26.23 &        0.929 &        0.0361 &        0.882 &        25.50 &        32.57 &        1.057 &        0.0568 &        0.777 &        22.33 &        64.23 &        2.863 &        0.0454 &        0.812 &        22.60 &        40.17 &        2.123 \\
    OSFV~\cite{wang2021one}              &        0.0403 &        0.865 &        25.66 &        30.21 &        1.279 &        0.0432 &        0.837 &        23.59 &        35.12 &        3.100 &        0.0589 &        0.746 &        21.56 &        67.40 &        2.432 &        0.0491 &        0.804 &        21.79 &        41.95 &        1.730 \\
    TPSMM~\cite{zhao2022thin}            &        0.0318 &        0.902 &        26.88 &        24.39 &        0.709 &        0.0359 &        0.886 &        25.53 &        32.77 &        0.983 &        0.0615 &        0.757 &        22.05 &        64.89 &        3.714 &        0.0516 &        0.780 &        22.10 &        40.84 &        2.254 \\
    LIA~\cite{wang2022latent}            &        0.0538 &        0.846 &        22.29 &        30.23 &        1.049 &        0.0477 &        0.879 &        24.43 &        38.89 &        0.932 &        0.0654 &        0.754 &        20.75 &        65.15 &        2.287 &        0.0537 &        0.815 &        21.47 &        42.27 &        1.502 \\
    DaGAN~\cite{hong2022depth}           &        0.0359 &        0.881 &        25.64 &        24.92 &        0.844 &        0.0413 &        0.846 &        23.95 &        34.35 &        2.405 &        0.0637 &        0.739 &        21.32 &        68.04 &        4.800 &        0.0453 &        0.826 &        22.56 &        37.36 &        1.523 \\
    Face2Face$^\rho$~\cite{yang2022face} &        0.0507 &        0.816 &        20.83 &        31.71 &        1.332 &        0.0466 &        0.832 &        22.45 &        37.64 &        1.772 &        0.0709 &        0.710 &        19.94 &        71.87 &        3.754 &        0.0649 &        0.764 &        19.55 &        84.57 &        1.863 \\
    \textbf{PECHead}                     &\textbf{0.0304}&\textbf{0.905}&\textbf{26.96}&\textbf{23.05}&\textbf{0.626}&\textbf{0.0357}&\textbf{0.903}&\textbf{26.76}&\textbf{30.10}&\textbf{0.746}&\textbf{0.0552}&\textbf{0.803}&\textbf{24.29}&\textbf{56.68}&\textbf{1.215}&\textbf{0.0435}&\textbf{0.859}&\textbf{23.03}&\textbf{31.20}&\textbf{0.839}\\
    \bottomrule
    \end{tabular}}%
    \vspace{-3mm}
\end{table*}

%% file: subsections/experiments.tex
\section{Experiments}
\label{sec:exp}

\input{sub_tex/fig_reenact}

\noindent\textbf{Datasets.}~We evaluate our model on VoxCeleb2~\cite{chung2018voxceleb2}, TalkingHead-1KH~\cite{wang2021one}, CelebV-HQ~\cite{zhu2022celebv}, and VFHQ~\cite{xie2022vfhq}.

\noindent\textbf{Implementation Details.}~In the generator $G$, the encoder and decoder are composed of two downsample and upsample ResBlocks~\cite{he2016deep}.
The Estimator $E$ consists of four downsample and upsample AdaIN~\cite{huang2017arbitrary} based ResBlocks.
The Reconstructor $R$ is separately trained and the landmarks obtained by the widely used framework~\cite{bulat2017far}.
More details about the datasets, network structures, and settings are provided in the supplementary material.

\noindent\textbf{Baselines.}~We compare our approach with the recently proposed representative methods, FOMM~\cite{siarohin2019first},  MRAA~\cite{siarohin2021motion}, OSFV~\cite{wang2021one}, TPSMM~\cite{zhao2022thin}, LIA~\cite{wang2022latent}, Face2Face$^\rho$~\cite{yang2022face} and DaGAN~\cite{hong2022depth}.
Our re-implementation of OSFV~\cite{wang2021one} is used with all settings followed by the original paper, while all other methods use the official implementation.

\noindent\textbf{Metrics.}~We use $L_1$, MS-SSIM~\cite{wang2004image, wang2003multiscale}, and PSNR to evaluate the low-level similarity between the synthesis and the driving images.
We also leverage FID~\cite{heusel2017gans} and FVD~\cite{unterthiner2018towards} to assess the image and video quality.
The average keypoint distance (AKD)~\cite{siarohin2019animating, siarohin2019first} is adopted to measure the semantic consistency.
The cross-identity similarity (CSIM)~\cite{wang2021one} is used to evaluate the identity preservation for cross-identity video face reenactment.
The average rotation distance (ARD)~\cite{doukas2021headgan} and the facial action unit hamming distance (AUH)~\cite{doukas2021headgan} are to measure errors of head pose angles and facial expressions.
For MS-SSIM, PSNR, and CSIM, larger values indicate better results, others the opposite.

\subsection{Same-identity Video Reconstruction}
\label{sec:sivr}

\input{sub_tex/tab_reenact}
We compare our models with state-of-the-art techniques for self-reenactment, where the source and driving frames depict the same individual. Quantitative results are presented in Tab.~\ref{tab:sivr}, and qualitative results are shown in Fig.~\ref{fig:sivr}. Our models demonstrate significant improvements across all metrics.
Most existing methods can generate satisfactory results for small pose movements, but for scenarios with significant pose variations, keypoint-based methods (FOMM~\cite{siarohin2019first}, MRAA~\cite{siarohin2021motion}, TPSMM~\cite{zhao2022thin}) may produce distorted faces due to a lack of 3D facial constraints.
The OSFV method~\cite{wang2021one}, which uses 3D keypoints, can produce consistent facial shapes, but the image quality is still unsatisfactory.
The depth-based method DaGAN~\cite{hong2022depth} has face distortion issues, indicating that self-supervised depth estimation is insufficient.
Latent vector-based models (LIA~\cite{wang2022latent}) perform poorly in capturing facial details and may entangle appearance information in the latent code.
The Face2Face$^\rho$ method~\cite{yang2022face} performs poorly due to inaccurate motion estimation.
Our method excels in preserving facial shape while accurately transferring facial expressions compared to existing techniques.

\input{sub_tex/tab_pose_exp}

\subsection{Cross-identity Video Face Reenactment}
\label{sec:civr}

We conducted experiments on the TalkingHead-1KH~\cite{wang2021one} and VFHQ~\cite{xie2022vfhq} datasets to explore cross-identity motion transfer, where the source and driving frames depict different individuals.
As shown in Fig.~\ref{fig:civr}, our method can produce convincing cross-identity face reenactment results that are more realistic, particularly in terms of facial expressions, compared to other techniques.
Keypoint-based methods (FOMM~\cite{siarohin2019first}, MRAA~\cite{siarohin2021motion}, TPSMM~\cite{zhao2022thin}) struggle to produce convincing results with noticeable face distortion for samples with large pose variations.
Face2Face$^\rho$~\cite{yang2022face} performs poorly since the landmarks only represent facial parts, making it challenging to handle non-facial characteristics like hair.
Quantitative results are presented in Tab.~\ref{tab:civr}.
Our method outperforms other techniques with the highest identity preservation ability and video quality, as well as the lowest pose angle and expression error.
Supplementary materials provide additional results, as well as subsequent experiments.

\input{sub_tex/fig_pose_exp}

\input{sub_tex/fig_wild}

\subsection{Head Pose and Expression Editing}
\label{sec:hpe}

For head pose editing, we compare the performance of frontalization.
The metric average rotation error (ARE)~\cite{doukas2021headgan} is adopted to measure the ability to control head pose, and FID is used to measure image quality.
For expression editing, we compare the performance of expression transfer, and AUH is used to measure the expression error.
Among the baselines, only the OSFV~\cite{wang2021one} and Face2Face$^\rho$ can manipulate the source frame without an explicit driving frame.
The results are shown in Tab.~\ref{tab:hpe} and Fig.~\ref{fig:hpe}.
Although the OSFV~\cite{wang2021one} has slightly better FID scores, it can not manipulate the pose and expression well.
The Face2Face$^\rho$~\cite{wang2021one} fails to estimate the flow field, causing poor results.

\input{sub_tex/fig_ablation}

\subsection{Free Editing on Wild Identities}
\label{sec:free_editing}
Finally, we demonstrate the strong capability of PECHead by editing wild face images downloaded from the Internet.
The results are shown in Fig.~\ref{fig:civr_free_head_exp}.
Our method can generate the face with desired head poses and expressions by changing the values of head pose $p$ and expression $e$.

\subsection{Ablation Studies}
\label{sec:ablation}

\input{sub_tex/tab_ablation}

To validate the effectiveness of each component, we first evaluate the performance of using both self-supervised learned and predefined facial landmarks.
We then assess the performance of the proposed MMFA module.
Finally, we evaluate the performance of the proposed video-based framework involving the CAP module.
Hence, we have the following settings:
(1) Self-supervised learned landmarks only (KP);
(2) Predefined landmarks only (LMK);
(3) Directly merge the learned and predefined landmarks (Direct);
(4) Concatenate the feature maps at a single level (FeatCat);
(5) The Motion-Aware Multi-Scale Feature Alignment (MMFA);
(6) The Full model.
The results in Fig.~\ref{fig:ablation} and Tab.~\ref{tab:ablation} reveal three important conclusions.
Firstly, utilizing both self-supervised learned landmarks and predefined landmarks is crucial to avoid face distortion and obtain high-quality results.
Secondly, the motion-aware multi-scale feature alignment (MMFA) module effectively aligns features from different scales, resulting in high-quality outcomes.
Lastly, the context adaptation and propagation (CAP) module propagates context information across frames, improving the smoothness of video synthesis.
Notably, only our full model produces high-fidelity results.

%% file: sub_tex/fig_reenact.tex
\begin{figure*}
    \captionsetup[subfigure]{aboveskip=1pt} %
    \captionsetup[subfigure]{labelformat=empty}
    \captionsetup[subfigure]{font={scriptsize}}
    \centering
    \begin{subfigure}[t]{0.098\linewidth}
        \includegraphics[width=\linewidth]
        {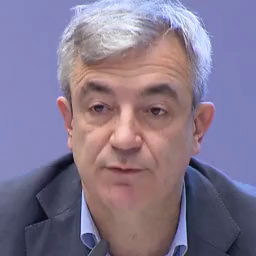} \\
        \vspace{-4.6mm}
        \includegraphics[width=\linewidth]
        {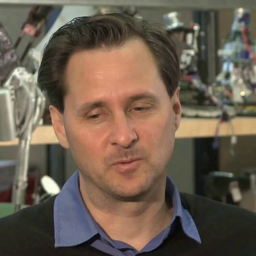} \\
        \vspace{-4.6mm}
        \includegraphics[width=\linewidth]
        {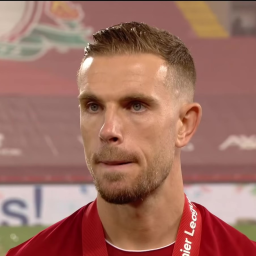} \\
        \vspace{-4.6mm}
        \includegraphics[width=\linewidth]
        {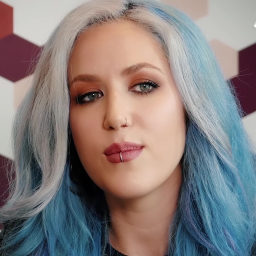} \\
        \vspace{-4.6mm}
        \includegraphics[width=\linewidth]
        {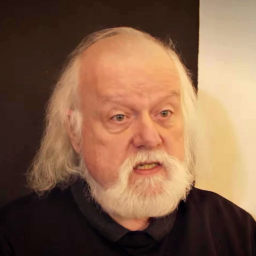}
        \caption{Source}
    \end{subfigure}%
    \begin{subfigure}[t]{0.098\linewidth}
        \includegraphics[width=\linewidth]
        {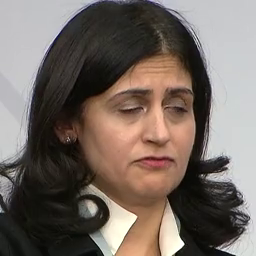} \\
        \vspace{-4.6mm}
        \includegraphics[width=\linewidth]
        {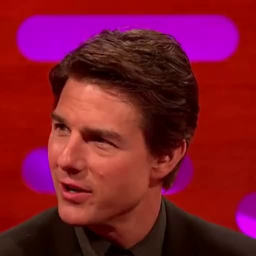} \\
        \vspace{-4.6mm}
        \includegraphics[width=\linewidth]
        {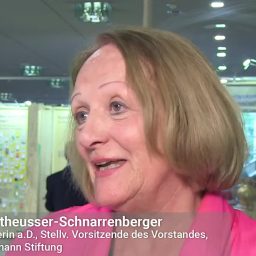} \\
        \vspace{-4.6mm}
        \includegraphics[width=\linewidth]
        {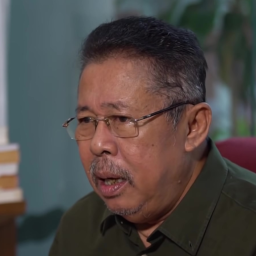} \\
        \vspace{-4.6mm}
        \includegraphics[width=\linewidth]
        {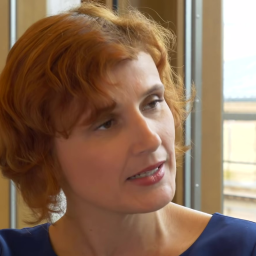}
        \caption{Driving}
    \end{subfigure}\hspace{1mm}%
    \begin{subfigure}[t]{0.098\linewidth}
        \includegraphics[width=\linewidth]
        {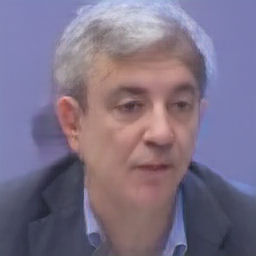} \\
        \vspace{-4.6mm}
        \includegraphics[width=\linewidth]
        {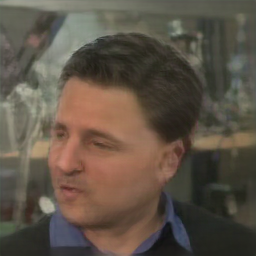} \\
        \vspace{-4.6mm}
        \includegraphics[width=\linewidth]
        {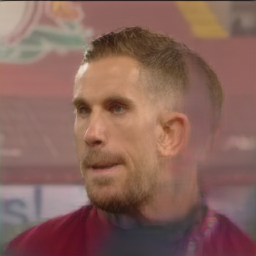} \\
        \vspace{-4.6mm}
        \includegraphics[width=\linewidth]
        {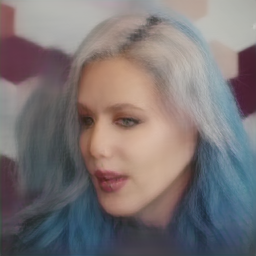} \\
        \vspace{-4.6mm}
        \includegraphics[width=\linewidth]
        {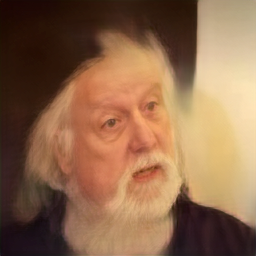}
        \caption{FOMM~\cite{siarohin2019first}}
    \end{subfigure}%
    \begin{subfigure}[t]{0.098\linewidth}
        \includegraphics[width=\linewidth]
        {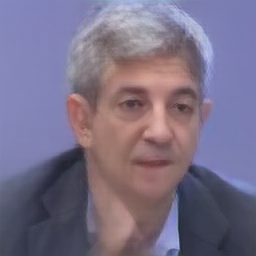} \\
        \vspace{-4.6mm}
        \includegraphics[width=\linewidth]
        {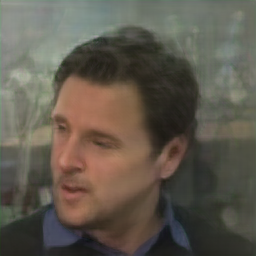} \\
        \vspace{-4.6mm}
        \includegraphics[width=\linewidth]
        {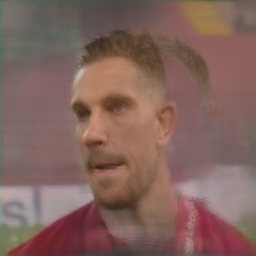} \\
        \vspace{-4.6mm}
        \includegraphics[width=\linewidth]
        {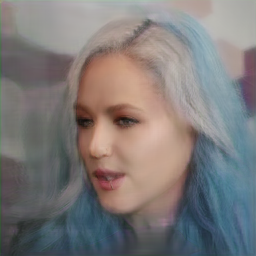} \\
        \vspace{-4.6mm}
        \includegraphics[width=\linewidth]
        {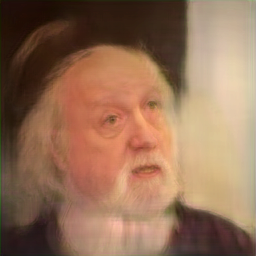}
        \caption{MRAA~\cite{siarohin2021motion}}
    \end{subfigure}%
    \begin{subfigure}[t]{0.098\linewidth}
        \includegraphics[width=\linewidth]
        {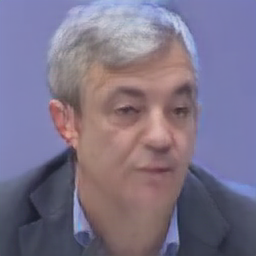} \\
        \vspace{-4.6mm}
        \includegraphics[width=\linewidth]
        {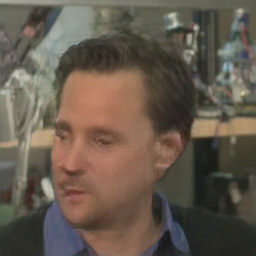} \\
        \vspace{-4.6mm}
        \includegraphics[width=\linewidth]
        {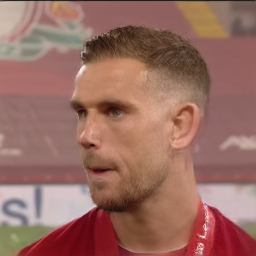} \\
        \vspace{-4.6mm}
        \includegraphics[width=\linewidth]
        {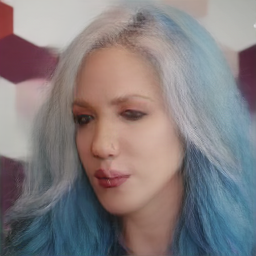} \\
        \vspace{-4.6mm}
        \includegraphics[width=\linewidth]
        {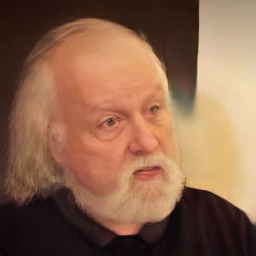}
        \caption{OSFV~\cite{wang2021one}}
    \end{subfigure}%
    \begin{subfigure}[t]{0.098\linewidth}
        \includegraphics[width=\linewidth]
        {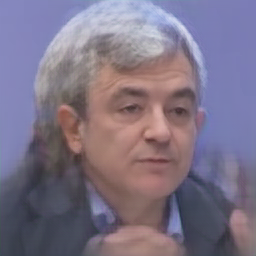} \\
        \vspace{-4.6mm}
        \includegraphics[width=\linewidth]
        {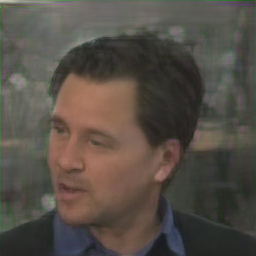} \\
        \vspace{-4.6mm}
        \includegraphics[width=\linewidth]
        {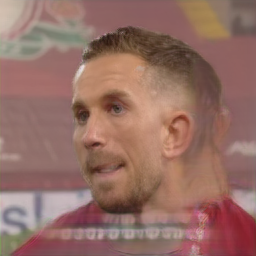} \\
        \vspace{-4.6mm}
        \includegraphics[width=\linewidth]
        {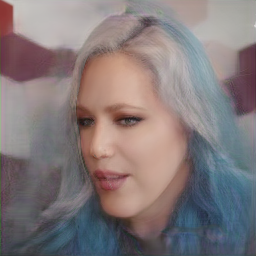} \\
        \vspace{-4.6mm}
        \includegraphics[width=\linewidth]
        {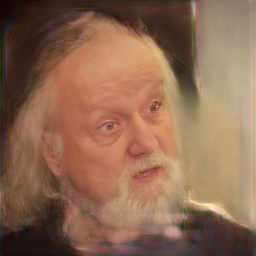}
        \caption{TPSMM~\cite{zhao2022thin}}
    \end{subfigure}%
    \begin{subfigure}[t]{0.098\linewidth}
        \includegraphics[width=\linewidth]
        {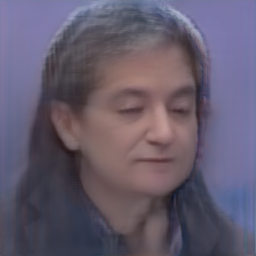} \\
        \vspace{-4.6mm}
        \includegraphics[width=\linewidth]
        {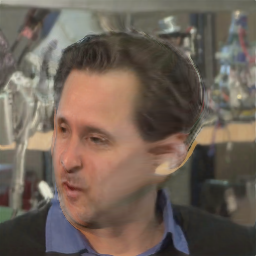} \\
        \vspace{-4.6mm}
        \includegraphics[width=\linewidth]
        {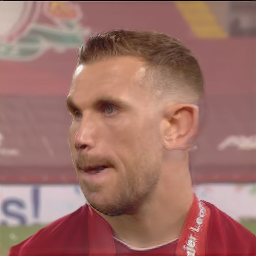} \\
        \vspace{-4.6mm}
        \includegraphics[width=\linewidth]
        {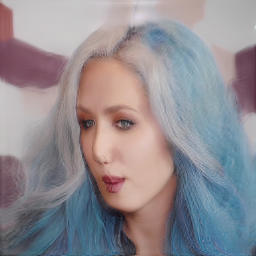} \\
        \vspace{-4.6mm}
        \includegraphics[width=\linewidth]
        {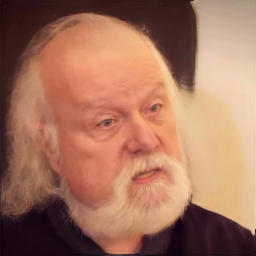}
        \caption{LIA~\cite{wang2022latent}}
    \end{subfigure}%
    \begin{subfigure}[t]{0.098\linewidth}
        \includegraphics[width=\linewidth]
        {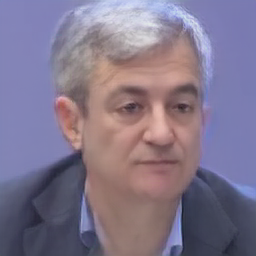} \\
        \vspace{-4.6mm}
        \includegraphics[width=\linewidth]
        {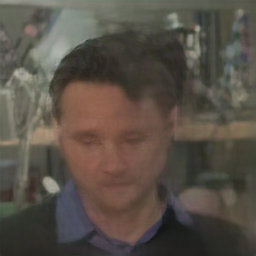} \\
        \vspace{-4.6mm}
        \includegraphics[width=\linewidth]
        {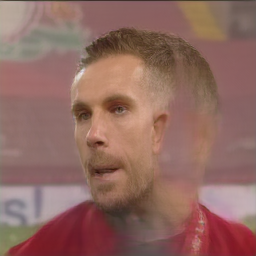} \\
        \vspace{-4.6mm}
        \includegraphics[width=\linewidth]
        {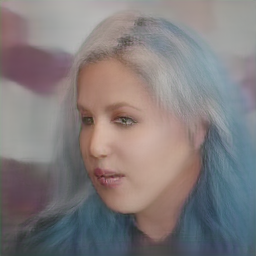} \\
        \vspace{-4.6mm}
        \includegraphics[width=\linewidth]
        {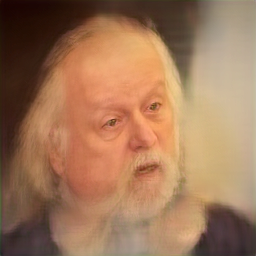}
        \caption{DaGAN~\cite{hong2022depth}}
    \end{subfigure}%
    \begin{subfigure}[t]{0.098\linewidth}
        \includegraphics[width=\linewidth]
        {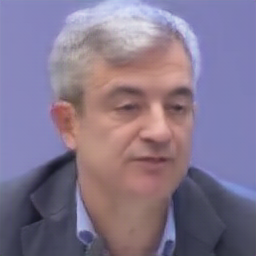} \\
        \vspace{-4.6mm}
        \includegraphics[width=\linewidth]
        {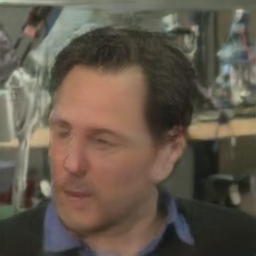} \\
        \vspace{-4.6mm}
        \includegraphics[width=\linewidth]
        {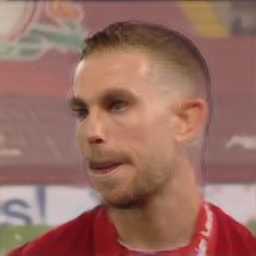} \\ %
        \vspace{-4.6mm}
        \includegraphics[width=\linewidth]
        {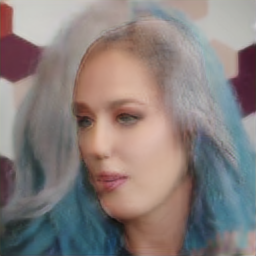} \\
        \vspace{-4.6mm}
        \includegraphics[width=\linewidth]
        {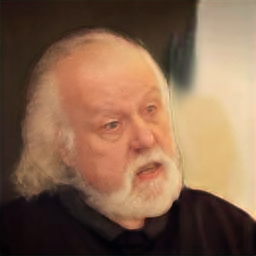}
        \caption{Face2Face$^\rho$~\cite{yang2022face}}
    \end{subfigure}%
    \begin{subfigure}[t]{0.098\linewidth}
        \includegraphics[width=\linewidth]
        {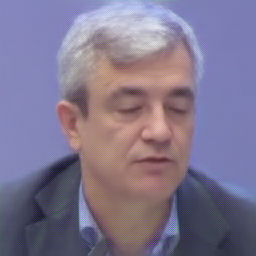} \\
        \vspace{-4.6mm}
        \includegraphics[width=\linewidth]
        {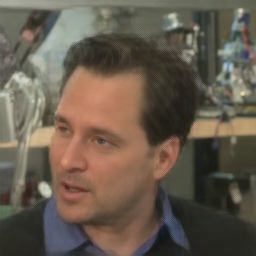} \\
        \vspace{-4.6mm}
        \includegraphics[width=\linewidth]
        {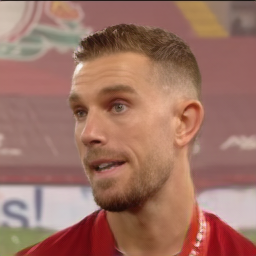} \\
        \vspace{-4.6mm}
        \includegraphics[width=\linewidth]
        {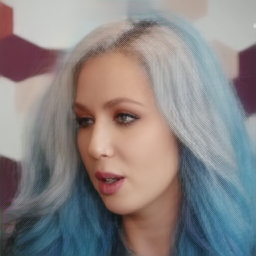} \\
        \vspace{-4.6mm}
        \includegraphics[width=\linewidth]
        {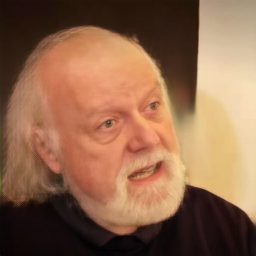}
        \caption{\textbf{PECHead}}
    \end{subfigure}
    \vspace{-2mm}
    \caption{Comparison of cross-identity face reenactment results obtained by the proposed method and other approaches.}
    \label{fig:civr}
    \vspace{-3.5mm}
\end{figure*}

%% file: sub_tex/tab_reenact.tex
\begin{table}[t]
    \caption{Quantitative results for the cross-identity reenactment.}
    \label{tab:civr}
    \vspace{-3mm}
    \centering
    \resizebox{\linewidth}{!}{%
    \begin{tabular}{l|cccc|cccc}
    \toprule
    \multirow{2}{*}{Methods} & \multicolumn{4}{c|}{CelebV-HQ} & \multicolumn{4}{c}{VFHQ} \\
                                         &         CSIM &         ARD &          AUH &          FVD &        CSIM  &         ARD &          AUH &         FVD   \\
    \midrule
    FOMM~\cite{siarohin2019first}        &        0.687 &        2.76 &        0.174 &        202.5 &        0.675 &        2.18 &        0.174 &        211.7  \\
    MRAA~\cite{siarohin2021motion}       &        0.670 &        2.65 &        0.145 &        219.1 &        0.662 &        2.07 &        0.159 &        205.9  \\
    OSFV~\cite{wang2021one}              &        0.706 &        3.21 &        0.171 &        207.3 &        0.754 &        4.11 &        0.205 &        213.4  \\
    TPSMM~\cite{zhao2022thin}            &        0.673 &        1.85 &        0.125 &        220.2 &        0.674 &        1.84 &        0.143 &        207.8  \\
    LIA~\cite{wang2022latent}            &        0.713 &        2.68 &        0.143 &        199.5 &        0.712 &        2.48 &        0.170 &        213.8  \\
    DaGAN~\cite{hong2022depth}           &        0.716 &        2.66 &        0.154 &        205.9 &        0.684 &        1.91 &        0.143 &        217.6  \\
    Face2Face$^\rho$~\cite{yang2022face} &        0.535 &        9.91 &        0.251 &        232.5 &        0.673 &        2.13 &        0.170 &        206.4  \\
    \textbf{PECHead}                     &\textbf{0.733}&\textbf{0.85}&\textbf{0.118}&\textbf{192.2}&\textbf{0.789}&\textbf{0.81}&\textbf{0.104}&\textbf{201.6} \\
    \bottomrule
    \end{tabular}}
    \vspace{-2mm}
\end{table}

%% file: sub_tex/tab_pose_exp.tex
\begin{table}[t]
    \caption{Quantitative results of pose and expression editing.}
    \label{tab:hpe}
    \vspace{-3mm}
    \centering
    \footnotesize
    \resizebox{\linewidth}{!}{%
    \begin{tabular}{l|ccc|ccc}
    \toprule
    \multirow{2}{*}{Methods} & \multicolumn{3}{c|}{TalkHead-1KH} & \multicolumn{3}{c}{VFHQ} \\
                                         &ARE                   & FID &          AUH &         ARE &           FID   &         AUH  \\
    \midrule
    OSFV~\cite{wang2021one}              &        4.89 &\textbf{40.96}&        0.136 &        3.46 &\textbf{\; 53.21}&        0.158 \\
    Face2Face$^\rho$~\cite{yang2022face} &        2.44 &        88.71 &        0.121 &        2.11 &          125.72 &        0.141 \\
    \textbf{PECHead}                     &\textbf{1.15}&        42.04 &\textbf{0.075}&\textbf{0.93}&        \; 56.16 &\textbf{0.080}\\
    \bottomrule
    \end{tabular}}
    \vspace{-3mm}
\end{table}

%% file: sub_tex/fig_pose_exp.tex
\begin{figure}[t]
    \captionsetup[subfigure]{labelformat=empty}
    \captionsetup[subfigure]{aboveskip=1pt} %
    \centering
    \begin{subfigure}[t]{\dimexpr0.182\linewidth+10pt\relax}
        \makebox[10pt]{\raisebox{20pt}{\rotatebox[origin=c]{90}{\footnotesize{Input}}}}%
        \includegraphics[width=\dimexpr\linewidth-10pt\relax]
        {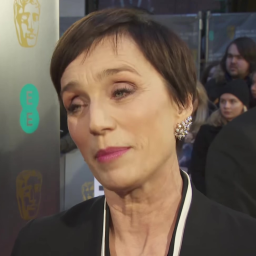}
        \makebox[10pt]{\raisebox{20pt}{\rotatebox[origin=c]{90}{\footnotesize{Face2Face$^\rho$}}}}%
        \includegraphics[width=\dimexpr\linewidth-10pt\relax]
        {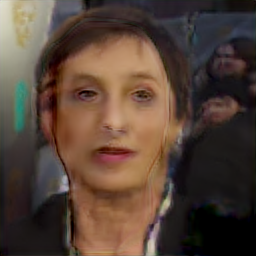}
        \makebox[10pt]{\raisebox{20pt}{\rotatebox[origin=c]{90}{\footnotesize{OSFV}}}}%
        \includegraphics[width=\dimexpr\linewidth-10pt\relax]
        {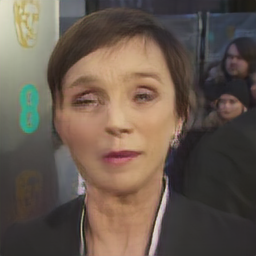}
        \makebox[10pt]{\raisebox{20pt}{\rotatebox[origin=c]{90}{\footnotesize{\textbf{PECHead}}}}}%
        \includegraphics[width=\dimexpr\linewidth-10pt\relax]
        {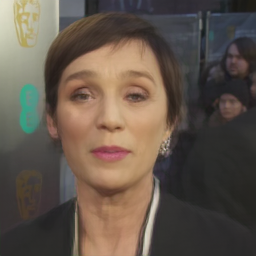}
    \end{subfigure}\hspace{0.4mm}%
    \begin{subfigure}[t]{0.182\linewidth}
        \includegraphics[width=\linewidth]
        {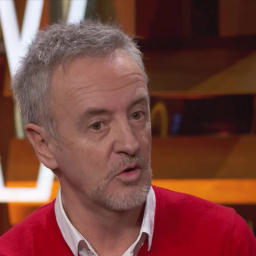}
        \includegraphics[width=\linewidth]
        {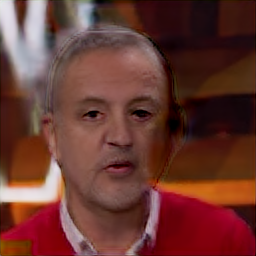}
        \includegraphics[width=\linewidth]
        {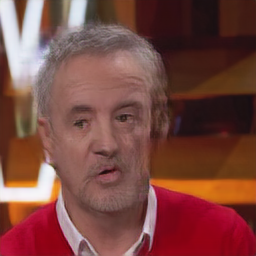}
        \includegraphics[width=\linewidth]
        {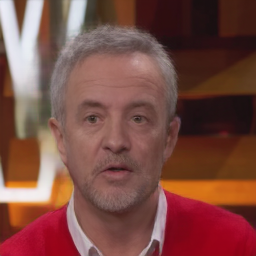}
    \end{subfigure}\hspace{0.4mm}%
    \begin{subfigure}[t]{0.182\linewidth}
        \includegraphics[width=\linewidth]
        {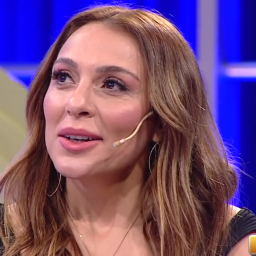}
        \includegraphics[width=\linewidth]
        {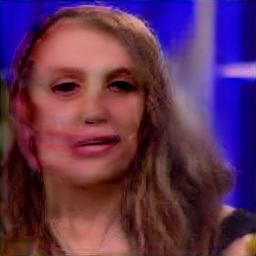}
        \includegraphics[width=\linewidth]
        {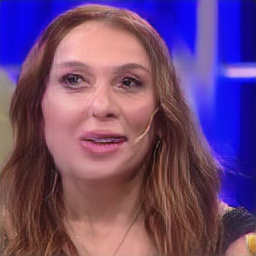}
        \includegraphics[width=\linewidth]
        {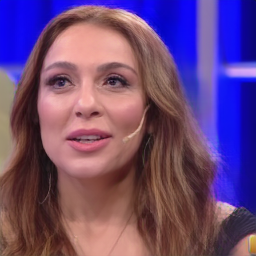}
    \end{subfigure}\hspace{0.4mm}%
    \begin{subfigure}[t]{0.182\linewidth}
        \includegraphics[width=\linewidth]
        {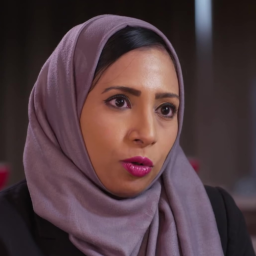}
        \includegraphics[width=\linewidth]
        {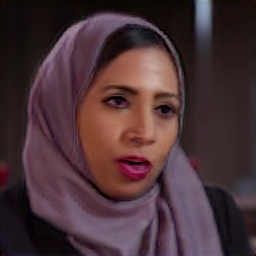}
        \includegraphics[width=\linewidth]
        {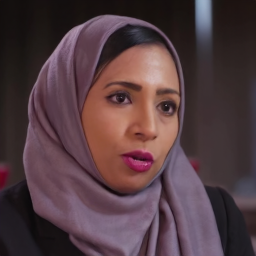}
        \includegraphics[width=\linewidth]
        {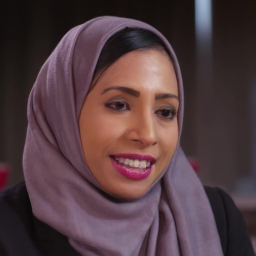}
    \end{subfigure}\hspace{0.4mm}%
    \begin{subfigure}[t]{0.182\linewidth}
        \includegraphics[width=\linewidth]
        {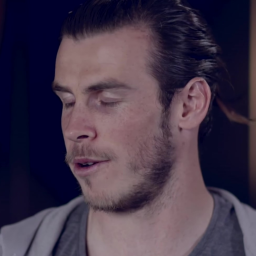}
        \includegraphics[width=\linewidth]
        {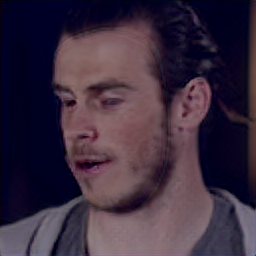}
        \includegraphics[width=\linewidth]
        {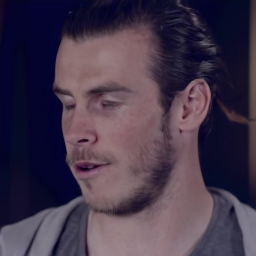}
        \includegraphics[width=\linewidth]
        {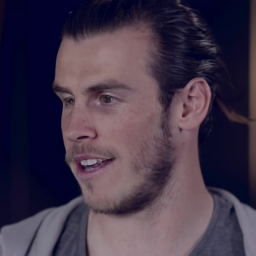}
    \end{subfigure}
    \vspace{-3mm}
    \caption{Head pose and expression editing results.}
    \label{fig:hpe}
    \vspace{-6mm}
\end{figure}

%% file: sub_tex/fig_wild.tex
\begin{figure*}[t]
    \captionsetup[subfigure]{labelformat=empty}
    \captionsetup[subfigure]{aboveskip=1pt} %
    \centering
    \begin{subfigure}[t]{0.134\linewidth}
        \includegraphics[width=\linewidth]
        {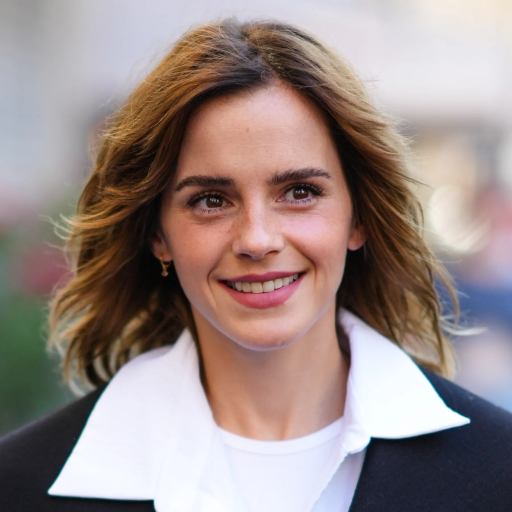}
        \includegraphics[width=\linewidth]
        {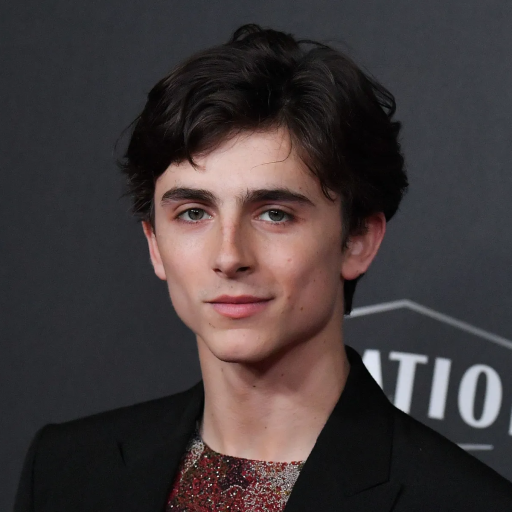}
        \caption{Input}
    \end{subfigure}\hspace{0.5mm}%
    \begin{subfigure}[t]{0.134\linewidth}
        \includegraphics[width=\linewidth]
        {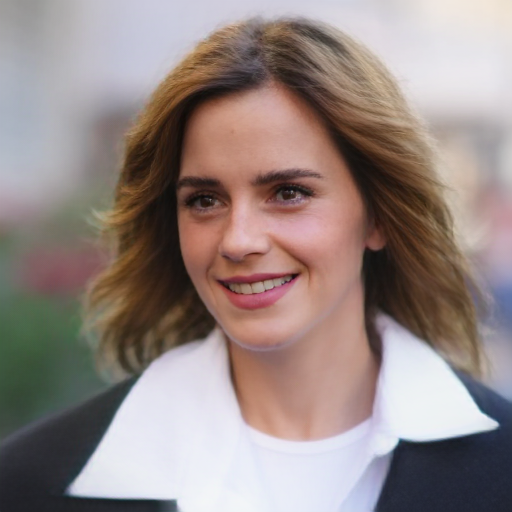}
        \includegraphics[width=\linewidth]
        {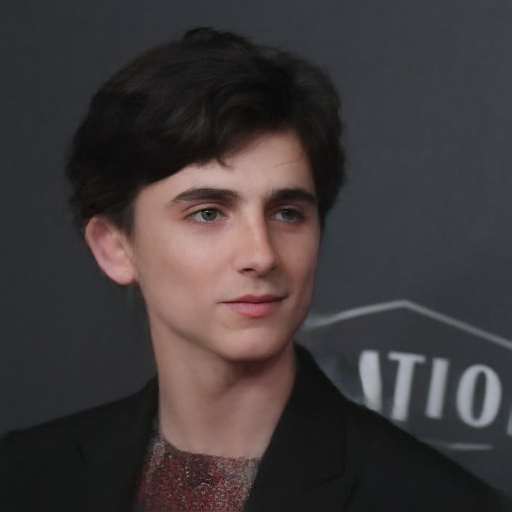}
        \caption{Yaw}
    \end{subfigure}\hspace{0.5mm}%
    \begin{subfigure}[t]{0.134\linewidth}
        \includegraphics[width=\linewidth]
        {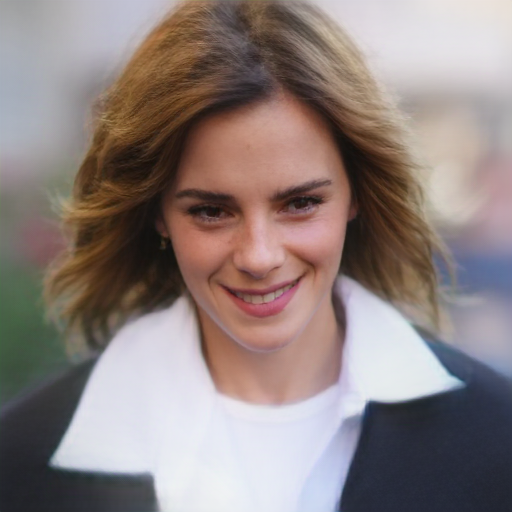}
        \includegraphics[width=\linewidth]
        {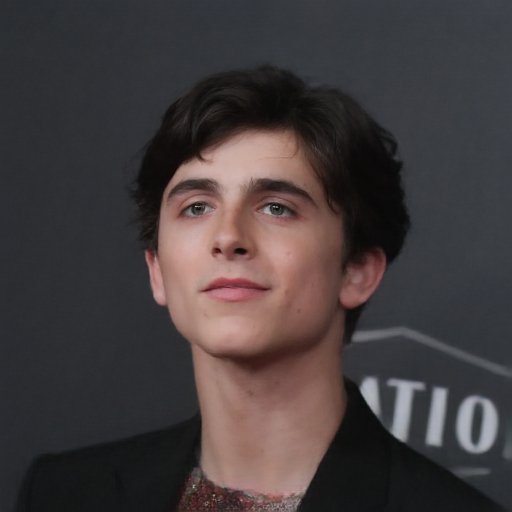}
        \caption{Pitch}
    \end{subfigure}\hspace{0.5mm}%
    \begin{subfigure}[t]{0.134\linewidth}
        \includegraphics[width=\linewidth]
        {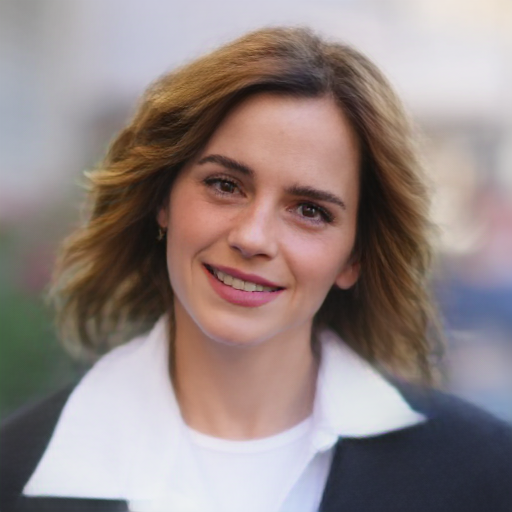}
        \includegraphics[width=\linewidth]
        {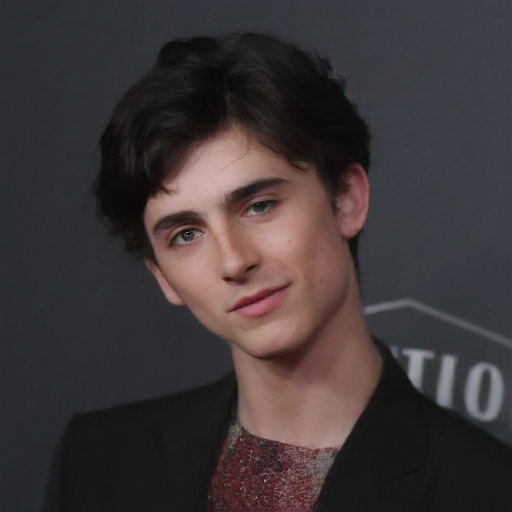}
        \caption{Roll}
    \end{subfigure}\hspace{0.5mm}%
    \begin{subfigure}[t]{0.134\linewidth}
        \includegraphics[width=\linewidth]
        {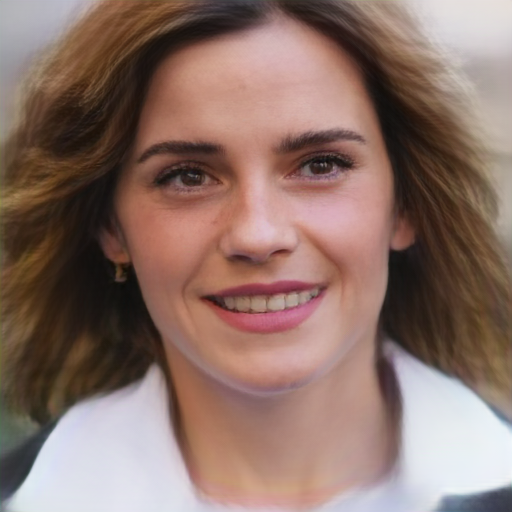}
        \includegraphics[width=\linewidth]
        {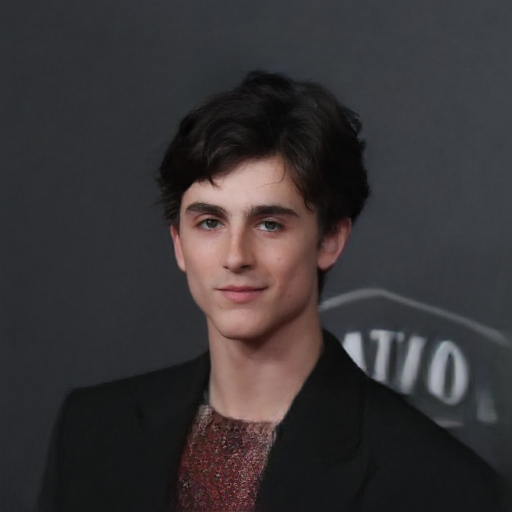}
        \caption{Field of view}
    \end{subfigure}\hspace{0.5mm}%
    \begin{subfigure}[t]{0.134\linewidth}
        \includegraphics[width=\linewidth]
        {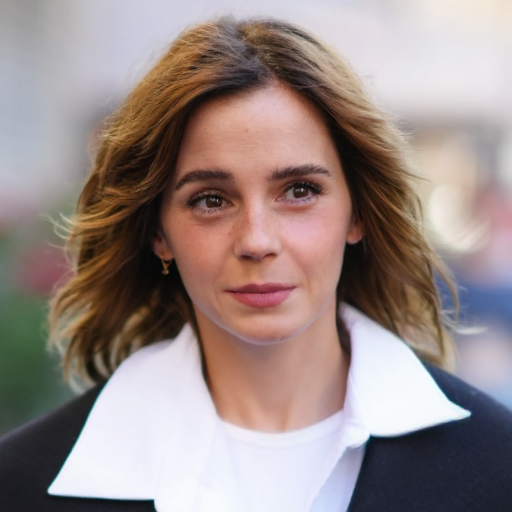}
        \includegraphics[width=\linewidth]
        {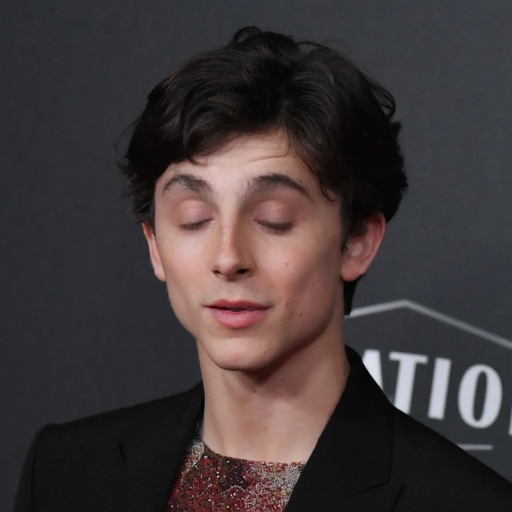}
        \caption{Expression}
    \end{subfigure}\hspace{0.5mm}%
    \begin{subfigure}[t]{0.134\linewidth}
        \includegraphics[width=\linewidth]
        {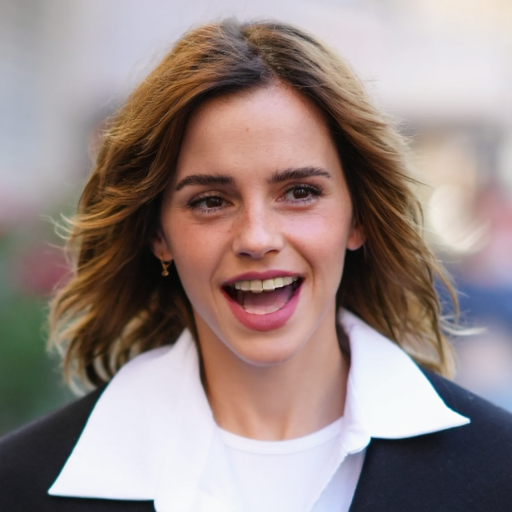}
        \includegraphics[width=\linewidth]
        {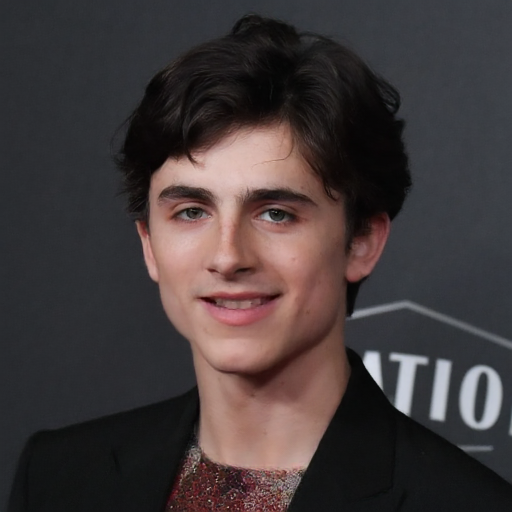}
        \caption{Expression}
    \end{subfigure}
    \vspace{-3mm}
    \caption{Head pose and expression freely controllable face reenactment.}
    \label{fig:civr_free_head_exp}
    \vspace{-3mm}
\end{figure*}

%% file: sub_tex/fig_ablation.tex
\begin{figure}[t]
    \captionsetup[subfigure]{labelformat=empty}
    \captionsetup[subfigure]{aboveskip=1pt} %
    \centering
    \begin{subfigure}[t]{0.21\linewidth}
        \includegraphics[width=\linewidth]
        {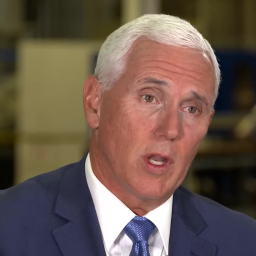}
        \caption{Source}
        \includegraphics[width=\linewidth]
        {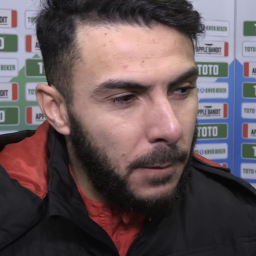}
        \caption{Driving}
    \end{subfigure}\hspace{1.5mm}%
    \begin{subfigure}[t]{0.21\linewidth}
        \includegraphics[width=\linewidth]
        {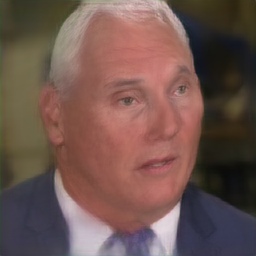}
        \caption{KP}
        \includegraphics[width=\linewidth]
        {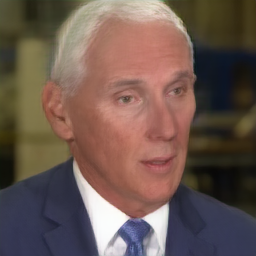}
        \caption{FeatCat}
    \end{subfigure}\hspace{1mm}%
    \begin{subfigure}[t]{0.21\linewidth}
        \includegraphics[width=\linewidth]
        {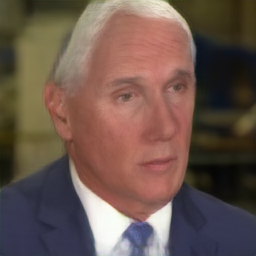}
        \caption{LMK}
        \includegraphics[width=\linewidth]
        {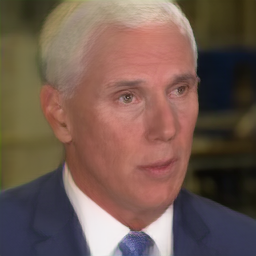}
        \caption{MMFA}
    \end{subfigure}\hspace{1mm}%
    \begin{subfigure}[t]{0.21\linewidth}
        \includegraphics[width=\linewidth]
        {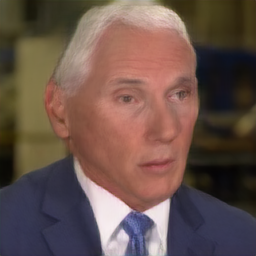}
        \caption{Direct}
        \includegraphics[width=\linewidth]
        {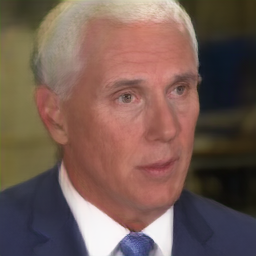}
        \caption{\textbf{Full}}
    \end{subfigure}%
    \vspace{-3mm}
    \caption{Ablation studies of our proposed model.}
    \label{fig:ablation}
    \vspace{-4mm}
\end{figure}

%% file: sub_tex/tab_ablation.tex
\begin{table}[t]
    \caption{Quantitative results for ablation studies.}
    \label{tab:ablation}
    \vspace{-3mm}
    \centering
    \resizebox{\linewidth}{!}{%
    \begin{tabular}{l|ccccc|ccccc}
    \toprule
    \multirow{2}{*}{Settings} & \multicolumn{5}{c|}{TalkHead-1KH} & \multicolumn{5}{c}{VFHQ} \\
                   & $L_1$  & FID & CSIM  & ARD & FVD  & $L_1$   & FID & CSIM  & ARD & FVD\\
    \midrule
    KP             &        0.0446 &        35.82 &        0.726 &        1.41 &        215.8 &        0.0491 &        37.8 &        0.712 &        1.40 &        218.5 \\
    LMK            &        0.0426 &        37.30 &        0.717 &        1.29 &        213.9 &        0.0485 &        36.6 &        0.709 &        1.37 &        217.9 \\
    Direct         &        0.0439 &        35.58 &        0.730 &        1.37 &        212.7 &        0.0474 &        32.9 &        0.724 &        1.33 &        217.8 \\
    FeatCat        &        0.0430 &        34.96 &        0.732 &        1.34 &        208.2 &        0.0462 &        32.0 &        0.733 &        1.09 &        213.9 \\
    MMFA           &        0.0375 &        31.27 &        0.764 &        0.81 &        206.8 &        0.0448 &\textbf{31.0}&        0.782 &        0.85 &        209.9 \\
    \textbf{Full}  &\textbf{0.0357}&\textbf{30.10}&\textbf{0.779}&\textbf{0.79}&\textbf{199.6}&\textbf{0.0435}&        31.2 &\textbf{0.789}&\textbf{0.84}&\textbf{201.6}\\
    \bottomrule
    \end{tabular}}
    \vspace{-4.5mm}
\end{table}

%% file: subsections/conclusion.tex
\section{Conclusion}
\label{sec:conclusion}

In this work, we present a novel method called PECHead, which generates high-fidelity talking head videos with free control over head pose and expression.
Leveraging both learned and predefined landmarks, we introduce a motion-aware multi-scale feature alignment module to model global and local movements simultaneously.
Furthermore, to improve the smoothness and naturalness of video synthesis, we introduce a context adaptation and propagation module that adapts the context of previous frames.
Our method outperforms existing approaches in face reenactment and controllable talking head generation, achieving state-of-the-art results.